\documentclass[runningheads]{llncs}

\usepackage{eccv}

\usepackage{eccvabbrv}

\usepackage{graphicx}
\usepackage{booktabs}
\usepackage[accsupp]{axessibility}  

\usepackage{diagbox}
\usepackage{mathtools}
\usepackage{multirow}
\usepackage{pifont}
\usepackage{siunitx}
\usepackage[table]{xcolor}

\newcommand{\dd}{\mathrm{d}}
\renewcommand{\vec}[1]{\mathbf{#1}}
\DeclarePairedDelimiter{\abs}{\lvert}{\rvert}

\newcommand{\VecWavePSF}{\vec{E}}
\newcommand{\VecWaveXP}{\vec{E}_t}

\newcommand{\NAAngle}{\Theta}
\newcommand{\NA}{\mathrm{NA}}
\newcommand{\DebyeSam}{N}
\newcommand{\DebyeSamBound}{N_\mathrm{inf}}
\newcommand{\IOR}{n_t}
\newcommand{\DeltaZ}{z}
\newcommand{\Wvl}{\lambda}
\newcommand{\Imag}{j}
\newcommand{\Foc}{f}

\newcommand{\InSam}{N}
\newcommand{\OutSam}{M}

\newcommand{\Ker}{P}
\newcommand{\KerSize}{k}
\newcommand{\Upsam}{u}

\newcommand{\SharpRAW}{I_s^{\text{RAW}}}
\newcommand{\SharpLIN}{I_s^{\text{lin}}}
\newcommand{\SharpRGB}{I_s^{\text{sRGB}}}

\newcommand{\BlurryLIN}{I_b^{\text{lin}}}
\newcommand{\BlurryRGB}{I_b^{\text{sRGB}}}

\newcommand{\Lens}{L}
\newcommand{\Focus}{f}

\DeclareMathOperator*{\argmin}{arg\,min}
\DeclarePairedDelimiter{\norm}{\lVert}{\rVert}

\usepackage{hyperref}
\usepackage{orcidlink}

\begin{document}

\title{Realistic Compound-Lens Defocus Blur Synthesis}
\titlerunning{Realistic Compound-Lens Defocus Blur Synthesis}

\author{Yunkyu Lee\inst{1}\orcidlink{0009-0005-5375-9330} \and
Woohyeok Kim\inst{1}\orcidlink{0009-0006-5691-5447} \and
Sunghyun Cho\inst{1}\orcidlink{0000-0001-7627-3513}}
\authorrunning{Y.~Lee et al.}
\institute{POSTECH, Pohang, Korea\\
\email{\{lyk1012,woohyeok,s.cho\}@postech.ac.kr}}

\maketitle

\begin{abstract}
Defocus blur degrades fine image structures and limits visual perception, which can adversely affect downstream vision tasks.
Although recent deep learning deblurring methods have achieved strong performance, their effectiveness depends on training data and often degrades across cameras and lenses due to limited optical diversity and realism in existing datasets.
In this paper, we propose a pipeline for synthesizing realistic defocus deblurring datasets for diverse compound lenses.
It integrates efficient wave-optics PSF computation via Debye CZT propagation, depth-aware defocus rendering with occlusion handling, and blur synthesis in the radiometrically linear space with camera ISP simulation.
This unified pipeline enables the scalable generation of photorealistic defocus datasets with diverse lens characteristics.
Using our pipeline, we generate CLDefocus, a large-scale synthetic dataset containing lens-diverse defocus image pairs.
We further analyze the limitations of real-captured defocus datasets and show that such imperfections can bias full-reference evaluation.
Extensive experiments demonstrate that models trained on CLDefocus achieve improved cross-device generalization compared to models trained on existing real and synthetic datasets.
Code and dataset are available at: \url{https://github.com/lykelee/CLDefocus}.
\end{abstract}

\section{Introduction}
\label{sec:intro}

Defocus blur arises when light rays from an object point fail to converge on the sensor plane, forming a blurred region known as the circle of confusion (CoC).
Optical systems employing large apertures often exhibit a shallow depth of field, which naturally leads to defocus blur.
Although such blur can be deliberately exploited for aesthetic purposes in photography, 
it is generally undesirable in computer vision, as it degrades fine details and hampers visual understanding~\cite{saadDefocusBlurInvariantScaleSpace2016}.
This degradation directly affects downstream tasks such as object detection, face recognition, and semantic segmentation~\cite{wangDualSuperResolutionLearning2020, kongFoveaBoxBeyoundAnchorBased2020}.

Recent advances in deep learning have led to significant progress in defocus deblurring~\cite{abuolaimDefocusDeblurringUsing2020, ruanAIFNetAllinFocusImage2021, leeIterativeFilterAdaptive2021, zamirRestormerEfficientTransformer2022, quanNeumannNetworkRecursive2023, quanSingleImageDefocus2023}, 
surpassing traditional blur-map-based deconvolution methods~\cite{shiDiscriminativeBlurDetection2014, krishnanFastImageDeconvolution2009}. 
However, the effectiveness of these methods is fundamentally constrained by the availability of realistic and diverse training datasets.
This dependency naturally raises the question of how defocus datasets are constructed, and what limitations they impose on generalization.

Real-captured defocus datasets are typically constructed by repeatedly capturing the same scene with different apertures~\cite{abuolaimDefocusDeblurringUsing2020, leeIterativeFilterAdaptive2021}.
This process scales poorly across cameras and lenses, limiting the diversity of optical characteristics and hindering cross-device generalization~\cite{yangEfficientDepthSpatiallyVarying2025}.
Moreover, changing optical settings often introduces brightness variations and spatial misalignment~\cite{leeIterativeFilterAdaptive2021, liLearningSingleImage2023}, and the resulting ground-truth images may still contain residual defocus blur or alignment errors.

Meanwhile, synthetic datasets have also been investigated. 
Early methods approximated defocus blur using disk or Gaussian kernels~\cite{potmesilSyntheticImageGeneration1982, dandresNonParametricBlurMap2016, shiJustNoticeableDefocus2015}, which fail to capture the lens-specific characteristics of real optics.
More recent works compute physically derived point spread functions (PSFs) via wave-optics simulations~\cite{chenOpticalAberrationsCorrection2021, luoCorrectingOpticalAberration2024, hoDifferentiableWaveOptics2025, mullerExaminingImpactOptical2026}.
However, these approaches often incur substantial computational cost and require empirically tuned sampling densities to avoid aliasing.
Such requirements hinder scalable PSF generation across diverse lens parameters, and most prior works therefore consider only a limited set of lens designs or fixed configurations rather than large-scale depth-varying defocus datasets.
Moreover, defocus blur is often synthesized directly in the non-linear sRGB color space without modeling the camera image signal processor (ISP)~\cite{rimRealisticBlurSynthesis2022}, reducing photometric realism.

In this paper, we propose a realistic compound lens defocus blur synthesis framework. 
It combines efficient and stable wave-optics PSF computation, depth-layered occlusion-aware defocus rendering, and linear-space blur synthesis with camera ISP simulation.
This unified pipeline enables scalable synthesis of photorealistic defocus datasets across diverse lenses and focus configurations.
It also avoids the alignment artifacts inherent in real-captured datasets while improving the photometric realism and computational efficiency of synthetic defocus generation.

Constructing such a realistic pipeline presents several practical challenges.
Wave-optics PSF computation under varying apertures and focus distances is computationally demanding and highly sensitive to sampling conditions~\cite{schmidtNumericalSimulationOptical2010}.
We address this by adopting a Debye CZT-based propagation framework~\cite{leuteneggerFastFocusField2006, huEfficientFullpathOptical2020} with explicit sampling criteria for stable and efficient PSF computation.
Rendering spatially varying defocus introduces another challenge because PSFs change continuously with scene depth, making direct per-pixel convolution with depth-varying PSFs computationally expensive.
To enable efficient rendering, we discretize depth using signed CoC values and perform layered compositing with occlusion handling~\cite{hasinoffLayerBasedRestorationFramework2007, krausDepthofFieldRenderingPyramidal2007}.
Photometric inconsistency also arises when blur is synthesized directly in the non-linear sRGB color space. 
To address this issue, we perform blur synthesis in the radiometrically linear color space with explicit ISP simulation rather than naive convolution in the sRGB space~\cite{rimRealisticBlurSynthesis2022}.

Using the proposed pipeline, we generate \emph{CLDefocus}, a large-scale synthetic dataset across diverse lens and focus configurations.
It consists of \num{40000} training pairs, \num{1000} validation pairs, and \num{1000} test pairs.
We also analyze the limitations of real defocus datasets and show that geometric and photometric inconsistencies in ground-truth pairs can bias full-reference evaluation, highlighting the need for physically grounded synthetic data.
Extensive experiments demonstrate that training on CLDefocus improves cross-device generalization compared to existing real and synthetic datasets.

Our contributions are summarized as follows:
\begin{itemize}
    \item We propose an efficient wave-optics PSF generation method for compound lenses based on the Debye CZT with explicit sampling criteria.
    
    \item We develop a realistic defocus rendering pipeline with depth-aware occlusion handling and blur synthesis in the radiometrically linear space with ISP-aware image formation.
    
    \item We present CLDefocus, a large dataset with lens-diverse synthetic defocus image pairs. We also analyze the limitations of real-captured defocus datasets and show that training on CLDefocus improves cross-device generalization across multiple real defocus benchmarks.
\end{itemize}
\section{Related Work}
\label{sec:related}

\subsection{Defocus Deblurring Methods}

Early works~\cite{bandoDigitalRefocusingSingle2007, zhuoDefocusMapEstimation2011, shiJustNoticeableDefocus2015, leeDeepDefocusMap2019} primarily focus on estimating defocus maps from images, which are then used with a conventional non-blind deconvolution method~\cite{krishnanFastImageDeconvolution2009}.
These methods depend on hand-crafted priors or simplified blur-kernel assumptions, limiting their robustness in complex real-world scenarios.
With the increasing availability of training data, learning-based methods have become dominant~\cite{ruanAIFNetAllinFocusImage2021, leeIterativeFilterAdaptive2021, abuolaimDefocusDeblurringUsing2020, sonSingleImageDefocus2021, maDefocusImageDeblurring2022, zamirRestormerEfficientTransformer2022, ruanLearningDeblurUsing2022, quanNeumannNetworkRecursive2023, quanSingleImageDefocus2023}.
End-to-end networks learn spatially varying blur patterns directly from data and significantly outperform traditional approaches.
However, their performance strongly depends on the diversity and realism of the training data, yet lens-diverse and physically accurate defocus datasets remain relatively limited.

\subsection{Real-Captured Defocus Datasets}

Acquiring high-quality blurred--sharp image pairs for defocus deblurring is inherently challenging, as it requires repeated captures of the same static scene under different aperture settings while keeping the camera pose fixed, limiting the scale and lens diversity of real-captured datasets. In practice, changing the aperture between captures often introduces brightness variations, spatial misalignment, and residual defocus even in the corresponding sharp images~\cite{leeIterativeFilterAdaptive2021, liLearningSingleImage2023}.

Despite these challenges, several real-captured defocus datasets have been proposed, but they remain limited in scale, camera diversity, or image formation. RTF~\cite{dandresNonParametricBlurMap2016} and RealDOF~\cite{leeIterativeFilterAdaptive2021} provide only test sets with a small number of scenes, and thus cannot be used to train learning-based defocus deblurring models.
DPDD~\cite{abuolaimDefocusDeblurringUsing2020} provides 500 blurred--sharp pairs captured by a Canon DSLR with dual-pixel (DP) sensors to obtain sub-aperture views and corresponding all-in-focus ground truth.
However, its scale remains limited and it is restricted to a single camera system. LFDOF~\cite{ruanAIFNetAllinFocusImage2021} increases dataset scale using light-field cameras, yet its image formation process differs from that of conventional cameras, potentially limiting generalization. SDD~\cite{liLearningSingleImage2023} captures data in dynamic driving scenarios, where blurred and sharp images are not perfectly aligned, requiring a dedicated training framework to handle the misalignment.

\subsection{Synthetic Defocus and Lens Blur Modeling}

Synthetic defocus generation has been explored as an alternative to real data. 
Early methods relied on simplified thin-lens models, approximating blur with disk or Gaussian kernels~\cite{bandoDigitalRefocusingSingle2007, shiJustNoticeableDefocus2015, leeDeepDefocusMap2019}.
While theoretically simple, these approximations fail to capture the lens-specific characteristics of real optical systems.

More physically grounded approaches compute PSFs using wave-optics simulations.
Some works adopt simple Fourier transform-based PSF formulations~\cite{mullerClassificationRobustnessCommon2023, chenPhysicsInformedBlurLearning2025}, while others employ more rigorous diffraction models such as the Huygens' principle~\cite{goodman2017introduction}, which is often formulated as the Rayleigh--Sommerfeld integral~\cite{chenOpticalAberrationsCorrection2021, luoCorrectingOpticalAberration2024, mullerExaminingImpactOptical2026}.
Although physically grounded and theoretically accurate, the Huygens' principle requires stringent sampling densities to avoid aliasing artifacts. 
In practice, increasing the defocus distance or aperture size necessitates finer sampling grids, leading to rapidly growing computational cost.
As a result, most prior studies restrict their evaluation to a limited number of lenses or primarily consider in-focus imaging, rather than systematically generating large-scale, depth-varying defocus datasets across diverse compound lenses.

In addition, many synthetic approaches perform blur rendering using simplified convolution models that do not fully account for occlusions in depth-of-field imaging or the camera imaging pipeline. 
ISP-aware modeling has been shown to be important for realistic blur synthesis~\cite{rimRealisticBlurSynthesis2022}, yet it is often neglected. 
In contrast, our work unifies physically grounded PSF computation across diverse compound lenses, depth-aware occlusion-consistent rendering, and ISP-aware image formation within a scalable dataset synthesis framework.
\section{PSF Computation}
\label{sec:psf}

Our realistic blur synthesis pipeline primarily relies on physically accurate point spread functions (PSFs) across diverse lenses, focus settings, scene depths, and field positions.
In this section, we describe how our pipeline enables efficient and stable computation of those PSFs.
Unlike thin-lens approximations based on geometric optics~\cite{leeDeepDefocusMap2019}, we model wave propagation through a compound lens.
To this end, we adopt a hybrid approach that combines ray tracing with wave-optics propagation~\cite{chenOpticalAberrationsCorrection2021, luoCorrectingOpticalAberration2024}.
Specifically, we acquire wavefront samples on the exit pupil sphere via ray tracing, fit them with Zernike polynomials~\cite{niuZernikePolynomialsTheir2022} to obtain a smooth analytical representation, and compute the sensor-plane wave field using the Debye formulation accelerated by the chirp Z-transform (CZT).
The resulting field yields lens-specific PSFs, enabling scalable generation across diverse lens configurations and defocus conditions.
We first introduce the Debye CZT for efficient PSF computation, followed by its application on compound lenses.

\subsection{Debye CZT}
\label{subsec:debye}

We introduce the diffraction model and numerical implementation used for PSF generation.
A defocused PSF is the intensity of the wave formed by a lens on a sensor plane displaced from the ideal focus.
This setting is naturally described by the Debye formulation~\cite{leuteneggerFastFocusField2006}, which models diffraction of converging light near focus.
We first present its Fourier form, and then discuss the sampling requirement and efficient implementation for a desired sensor region of interest (ROI).

\begin{figure*}[t]
  \centering
  \includegraphics[width=\linewidth]{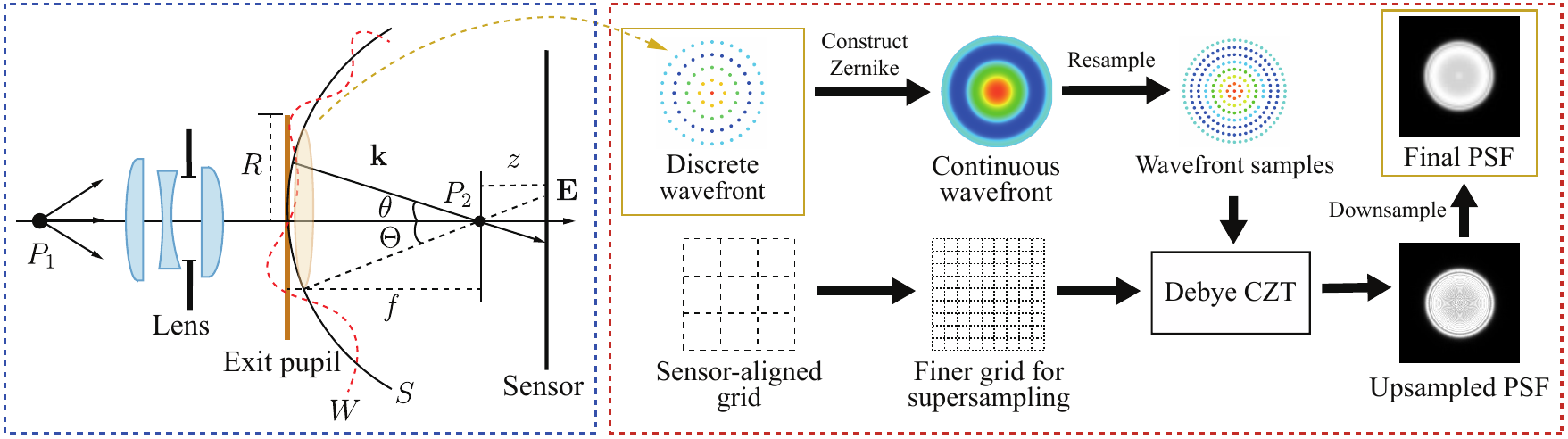}
  \caption{
    Overview of our PSF computation process.
    The left shows the physical setup of the Debye formulation.
    A point source $P_1$ emits light, which passes through the lens and forms a wavefront $W$.
    $S$ is the exit pupil sphere that is centered at a focal point $P_2$ and passes through the exit pupil center.
    We model $W$ as a converging spherical wave on $S$ with deviations (aberrations).
    The Debye formulation integrates all wave field within a cap with half-angle $\NAAngle$ to obtain the wave field on the image plane with defocus $z$.
    The right shows the PSF computation process with the CZT.
    We acquire discrete wavefront samples from the ray tracing result.
    Then we fit them with Zernike polynomials and resample a sample grid required by the CZT.
    We then perform the Debye CZT on a finer grid, obtaining a supersampled PSF. 
    Finally, we downsample it to obtain a PSF that matches the sensor resolution.
    }
  \label{fig:debye}
\end{figure*}

The Debye formulation expresses the propagated wave field $\VecWavePSF$ at an image point as the integral of the plane-wave spectrum over the exit pupil (XP) sphere, as illustrated on the left side of \cref{fig:debye}. For an on-axis point source, Leutenegger et al.~\cite{leuteneggerFastFocusField2006} rewrite the Debye formulation as a Fourier transform:
\begin{equation}\label{eqn:debye_ft}
	\VecWavePSF(x, y, z) = - \frac{\Imag f}{\Wvl k^2} \mathcal{F}\left\{ \VecWaveXP (\theta, \phi) \frac{\exp(\Imag k_z z)}{\cos \theta} \right\} (x, y, z),
\end{equation}
where $\VecWaveXP$ denotes the wave field on the XP sphere, $\Foc$ is the focal length, $\Wvl$ is the wavelength, $k$ is the wave number, $\DeltaZ$ is the axial defocus displacement, and $\Imag$ denotes the imaginary unit. The Fourier transform is with respect to the $x$, $y$ coordinates of the transverse wave vectors $\mathbf{k}$, which are proportional to the wave number and the normal vector on the XP sphere. This Fourier form allows the sensor-plane field to be computed in the Fourier domain after discretization.

For off-axis point sources, we adopt the approach proposed by Cai et al.~\cite{caiDirectCalculationTightly2019}. We first rotate the coordinate system to align the focal point with the new optical axis. Then we apply the on-axis Debye formulation with the image plane tilted with respect to the new coordinates. This extends \cref{eqn:debye_ft} and leads to a more involved formula. Details are provided in the supplementary material.

When the continuous Debye formulation is discretized onto a finite wave grid, a sufficient number of samples is required to prevent aliasing. Following the analysis in Leutenegger et al.~\cite{leuteneggerFastFocusField2006}, the sampling condition is given by:
\begin{equation}\label{eqn:debye_ft_sampling}
N > \DebyeSamBound, \quad
\DebyeSamBound = \frac{4 \NA^2}{\sqrt{\IOR^2 - \NA^2}} \frac{|\DeltaZ|}{\Wvl},
\end{equation}
where $N$ is the number of samples per coordinate, $\DebyeSamBound$ is the lower bound of $N$, $\NA$ denotes the numerical aperture, and $\IOR$ is the refractive index of the transmission medium. This condition originates from the phase factor $\exp(j k_z \DeltaZ)$ in \cref{eqn:debye_ft} and implies that higher NA and stronger defocus require more samples. It does not account for aliasing caused by rapid variations in $\VecWaveXP$, which can be non-negligible when lens aberrations are large. We therefore set $\DebyeSam = 2\DebyeSamBound$, which suppresses aliasing across all lenses and defocus levels considered.

When implementing the Fourier-based computation of the Debye formulation, a straightforward approach is to use the fast Fourier transform (FFT).
However, in the standard FFT, the output sampling pitch and ROI cannot be chosen freely, as the output grid is fixed by the input wavefront sampling
grid~\cite{schmidtNumericalSimulationOptical2010}. 
In contrast, our PSFs must be evaluated on a sensor-aligned grid whose sampling pitch
matches the sensor pixel pitch. To this end, we employ the chirp Z-transform (CZT)~\cite{leuteneggerFastFocusField2006, huEfficientFullpathOptical2020}
to flexibly control both the sampling resolution and the spatial extent for sampling.
The CZT allows flexible ROI selection and sampling resolution while maintaining computational efficiency via Bluestein's algorithm~\cite{bluesteinLinearFilteringApproach1970, rabinerChirpZtransformAlgorithm1969}.
We refer to this implementation as the \textbf{Debye CZT}.
Detailed derivations are provided in the supplementary material.

The Debye formulation is often written in vectorial form, where the wave field is vector-valued and polarization is explicitly modeled.
While this is important for high-NA optics and polarization-sensitive imaging, it is unnecessary in our setting.
We therefore use a scalar diffraction approximation by treating wave fields as scalar-valued. This choice is supported by two considerations. First, the photographic lenses used in our study have a low NA, as described in \cref{subsec:lens_collection}. Second, natural scenes typically exhibit negligible polarization effects~\cite{kupinskiAngleLinearPolarization2019}.

\subsection{Computing Lens PSFs}

While the Debye CZT has been used for microscopy with on-axis and analytically defined pupil functions~\cite{mioraCalculatingPointSpread2024}, existing literature provides limited guidance on how to apply it to photographic compound lenses. We use a three-stage procedure to compute depth- and field-dependent PSFs of a compound lens, given the focusing distance, object depth, field position, and sensor resolution. As illustrated on the right side of \cref{fig:debye}, the procedure consists of:
(1) wavefront sampling on the exit pupil sphere, where discrete wavefront samples are obtained from the lens prescription through ray tracing;
(2) Zernike-based wavefront representation, which reconstructs a smooth analytical wavefront from the samples; and (3) diffraction computation via the Debye CZT, which propagates the wavefront to the sensor plane and produces a PSF for the desired pixel pitch and region of interest. 
Further details are provided in the supplementary material.

\subsubsection{Wavefront Sampling on the Exit Pupil Sphere.}

For a given lens design, we perform dense ray tracing from a point source at the given depth and field through the lens.
We project all traced rays onto the exit pupil (XP) sphere and compute the optical path difference (OPD) for each ray.
The OPDs are then converted into phase delays, yielding discrete wave field samples.
Furthermore, to handle complicated pupil shapes of off-axis fields, we additionally calculate implicit function values to indicate whether a ray is blocked inside the lens.
They are used to reconstruct a ray clipping mask that determines the pupil shape.

\subsubsection{Zernike-Based Wavefront Representation.}

Ray tracing provides wavefront samples only at specific discrete positions, which do not generally form uniform grids that can be used for CZT.
Hence, we fit the sampled wavefront with Zernike polynomials to construct a continuous analytical representation.
We then resample the wavefront onto the sampling grid required by the CZT.
This model captures lens-specific aberrations and serves as the input field $\VecWaveXP$ in \cref{eqn:debye_ft}.

\subsubsection{Diffraction Computation via Debye CZT.}

We evaluate the sensor-plane wave field using the Debye formulation accelerated with the chirp Z-transform (CZT).
Let the desired PSF array be $\Ker \in \mathbb{R}^{\KerSize \times \KerSize}$ on the sensor grid.
Although the CZT allows direct sampling on this grid, insufficient output sampling may introduce aliasing artifacts.
To mitigate this, we upsample the output grid by a factor $\Upsam > 1$, compute the wave field $E_u$, and form the intensity $\Ker_u = |E_u|^2$.
The final PSF $P$ is obtained by downsampling $\Ker_u$ by a factor of $\Upsam$ to match the target resolution.
In practice, we use $\Upsam = 5$, which removes observable aliasing in all tested configurations.
\section{Defocus Dataset Synthesis}
\label{sec:synthesis}

\begin{figure*}[t]
    \centering
    \includegraphics[width=0.95\linewidth]{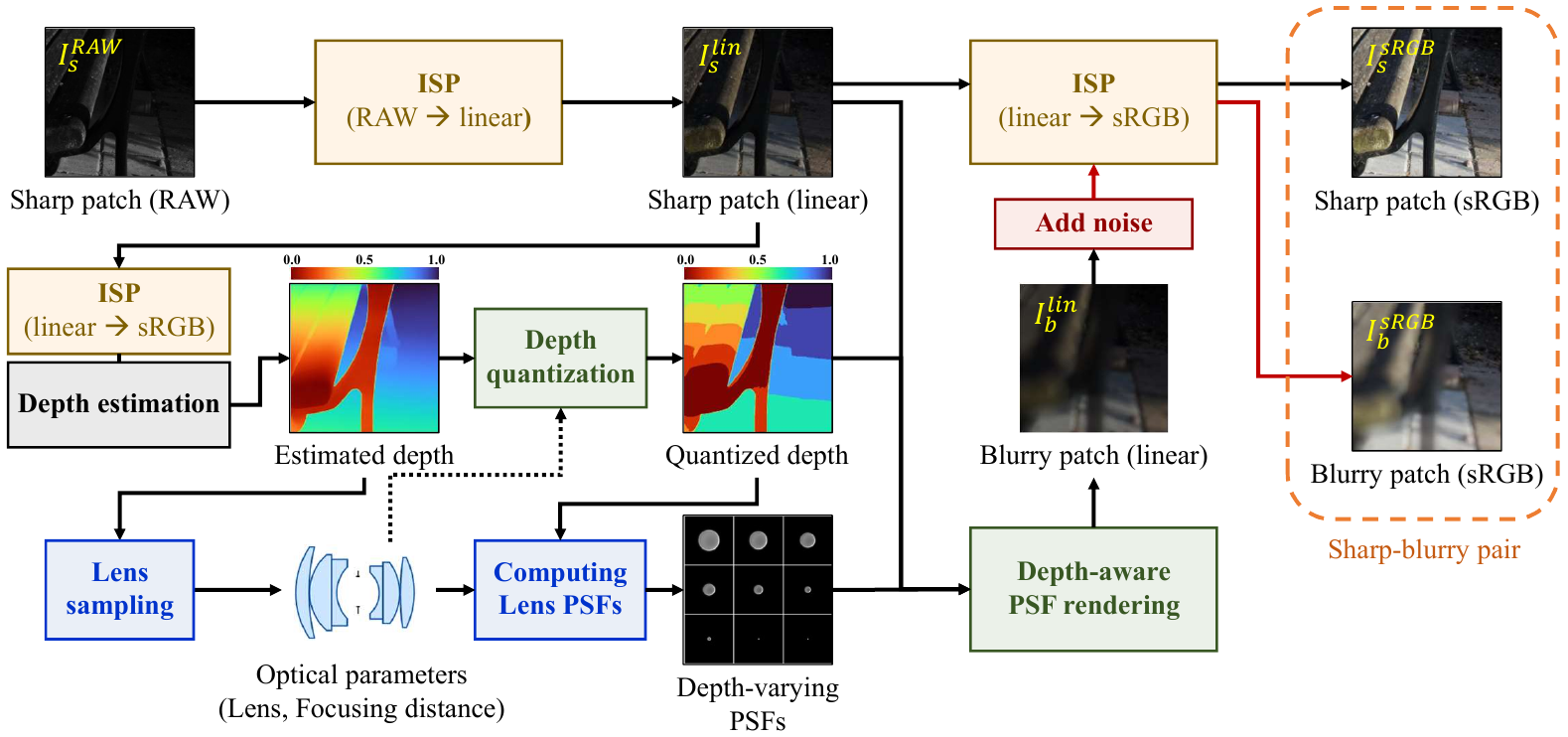}
    \caption{
    Overview of our data synthesis pipeline.
    From a sharp RAW image, we first sample a RAW patch and convert it to linear RGB via a partial ISP.
    Using the corresponding depth map, we sample a lens and a focusing distance that satisfy our constraints on blur size and practical PSF computation.
    We then quantize the depths while minimizing defocus discontinuity and build a multi-layered representation.
    Next, we compute PSFs for varying depths using the Debye CZT and perform depth-of-field imaging in the linear space, yielding the blurred patch.
    Finally, we synthesize noise and apply the ISP to obtain the blurred patch in the sRGB space.
    We also obtain the corresponding sharp patch by passing the linear sharp patch through the ISP.
    }
    \label{fig:main}
\end{figure*}

In this section, we present a realistic blur synthesis pipeline for generating defocused images from compound lens designs.
We synthesize depth-dependent defocus blur using the PSFs generated in \cref{sec:psf}, constructing physically consistent blurred--sharp training and test pairs from sharp RAW images through a depth-aware rendering pipeline (\cref{fig:main}).
Our synthesis operates across RAW, linear RGB, and non-linear sRGB color spaces, following the ISP pipeline of RSBlur~\cite{rimRealisticBlurSynthesis2022}.
Blur synthesis is performed in linear RGB, while the final pairs are produced in the non-linear sRGB color space.

\subsection{Lens Collection \& Filtering}
\label{subsec:lens_collection}

To generate diverse PSFs, we collect compound lens designs from the publicly available collection \cite{lensdesigns}.
We download Zemax files under ``Photographic primes'' and ``Photographic zooms'', retaining \num{1281} files after removing incompatible ones.
To reflect practical photographic optics, we filter the designs by basic properties such as F-number and total track length, keeping only lenses within typical photography ranges.
We further exclude lenses with $\NA \ge 0.6$, which fall outside standard photographic regimes and require excessive sampling under \cref{eqn:debye_ft_sampling}, as well as designs exhibiting severe aberrations that destabilize wavefront reconstruction and PSF computation.
After filtering, 700 lens designs remain for PSF generation.
The detailed criteria are provided in the supplementary material.

\subsection{Depth Estimation \& Lens Sampling}

Given a collection of sharp RAW source images, which we describe in \cref{sec:experiments}, we sample a RAW patch from each image as a sharp source patch.
The sampled RAW patch $\SharpRAW \in \mathbb{R}^{h \times w \times 1}$ is then converted to the linear space via the partial ISP, which applies demosaicing, white balancing, and color correction.
This produces the linear sharp patch
$\SharpLIN \in \mathbb{R}^{h \times w \times 3}$.
For training pairs, we also apply a few augmentations (rotation, flip, resizing, and exposure scaling) at this stage.
We obtain the sRGB patch $\SharpRGB$ via the forward ISP.
For each patch, we also obtain the corresponding depth map.
To obtain the depth map, we run a monocular estimator, Depth Pro~\cite{bochkovskiiDEPTHPROSHARP2025} on the full-resolution sharp sRGB image (obtained via the forward ISP), and crop the resulting metric depth map to match the linear sharp patch.

Next, we randomly sample a compound lens $\Lens$ and a focusing distance $\Focus$ for the patch.
To ensure feasible rendering, candidate lenses are evaluated using paraxial approximations. 
Lenses are discarded if the maximum circle of confusion (CoC) over the patch exceeds a predefined threshold (limiting PSF size) or if the required sampling number for the Debye CZT becomes excessive.
After this filtering, a valid $(\Lens,\Focus)$ pair is sampled from the remaining lenses.
The detailed sampling process and filtering criteria are provided in the supplementary material.

\subsection{Depth-Aware PSF Rendering}

Computing distinct PSFs for all continuous depth values is computationally infeasible.
We therefore quantize the depth map into $K$ discrete layers based on the \emph{signed} CoC, which distinguishes points in front of and behind the focal plane.
Depth values whose signed CoC difference is smaller than one sensor pixel are grouped into the same layer to reduce quantization artifacts.

This produces layered color images $\{C_i\}_{i=1}^{K}$ and opacity masks $\{A_i\}_{i=1}^{K}$:
\begin{equation}
C_i \in \mathbb{R}^{h \times w \times 3},
\quad
A_i \in \{0,1\}^{h \times w}.
\end{equation}
Here, $C_i$ is the color image of the $i$-th layer, and $A_i$ indicates its spatial support.
We note that regions occluded by foreground layers might contribute to the blurred result.
Hence, we inpaint those regions and set opacity masks to $1$.

Blur synthesis is performed entirely in the radiometrically linear space.
For each layer $i$, we retrieve the corresponding depth-dependent PSF $\Ker_i \in \mathbb{R}^{k \times k}$ and convolve it with $C_i$.
Since defocus with multiple depths cannot be expressed as a single spatially invariant convolution, we adopt the layered compositing formulation~\cite{hasinoffLayerBasedRestorationFramework2007, krausDepthofFieldRenderingPyramidal2007} and accumulate contributions from back to front:
\begin{equation}
    C^{\mathrm{blur}}_i 
    =
    \Ker_i \ast C_i 
    + 
    \left(1 - \Ker_i \ast A_i \right) 
    \cdot C^{\mathrm{blur}}_{i-1},
\end{equation}
where $\ast$ denotes convolution and 
$C^{\mathrm{blur}}_i \in \mathbb{R}^{h \times w \times 3}$ is the intermediate composite after processing layers $1,\ldots,i$.
The final blurred linear image is $\BlurryLIN = C^{\mathrm{blur}}_{K}$.

To ensure photometric consistency, $\BlurryLIN$ is converted to RAW space via the inverse partial ISP, sensor noise is synthesized, and the forward ISP is applied to obtain the blurred sRGB image $\BlurryRGB \in \mathbb{R}^{h \times w \times 3}$.
The sharp counterpart $\SharpRGB$ is generated by passing $\SharpLIN$ through the same ISP without degradation, yielding the blurred--sharp pair $(\BlurryRGB, \SharpRGB)$.
Random sampling of source images, augmentations, lenses, and focusings enables diverse training and testing pairs.
Additional implementation details are provided in the supplementary material.
\section{Experiments}
\label{sec:experiments}

\subsubsection{Implementation.}

For dataset synthesis, we implemented the PSF computation and the synthesis pipeline using JAX~\cite{jax2018github}, a Python package for GPU-accelerated numerical computations.
We also adopted ZERNIPAX~\cite{elmaciogluZERNIPAXFastAccurate2025} for efficient Zernike polynomial evaluation.
We primarily used NRKNet~\cite{quanNeumannNetworkRecursive2023} as a deblurring model.
We followed its original training protocols as much as possible.
For example, NRKNet was originally trained for \num{4000} epochs with batch size 4 for DPDD, which has 350 training pairs.
We used \( \num{350000} = \num{4000} \cdot 350 / 4 \) iterations to match the total iterations.
We observe the same trend for other deblurring models~\cite{zamirRestormerEfficientTransformer2022, quanSingleImageDefocus2023, chenSimpleBaselinesImage2022} and provide the corresponding results in the supplementary material.

\subsubsection{CLDefocus Dataset.}

We generate CLDefocus using our synthesis pipeline with train, validation, and test splits.
For fair evaluation, we use disjoint splits for both sharp source images and lenses.
As sharp sources, we employ sharp images of DPDD~\cite{abuolaimDefocusDeblurringUsing2020} in RAW format, following its original 350/74/76 split for train/validation/test.
The 700 lenses in \cref{subsec:lens_collection} are split into 420/140/140 lenses.
This yields \num{40000}/\num{1000}/\num{1000} train/validation/test pairs at $384 \times 384$ resolution.
To mitigate residual blur in the ground-truth images, we generate \SI{50}{\percent} extra patches and discard lower-quality pairs using sharpness filtering~\cite{abuolaimDefocusDeblurringUsing2020}.

\subsection{Comparison with Other Datasets}

\begin{table}[t]
	\centering
	\setlength{\tabcolsep}{1pt}
	\caption{Quantitative results on full-reference metrics (PSNR, SSIM, LPIPS~\cite{zhangUnreasonableEffectivenessDeep2018}). The best and second-best results are highlighted in bold and underlined, respectively.}
	\label{tab:result_main_fullref}
	\resizebox{1\linewidth}{!}{
		\begin{tabular}{lcccccccccccc}
			\toprule
			Test set & \multicolumn{3}{c}{CLDefocus} & \multicolumn{3}{c}{RTF} & \multicolumn{3}{c}{RealDOF} & \multicolumn{3}{c}{DPDD} \\
			\cmidrule(lr){2-4}\cmidrule(lr){5-7}\cmidrule(lr){8-10}\cmidrule(lr){11-13}
			Train set & PSNR$\uparrow$ & SSIM$\uparrow$ & LPIPS$\downarrow$ & PSNR$\uparrow$ & SSIM$\uparrow$ & LPIPS$\downarrow$ & PSNR$\uparrow$ & SSIM$\uparrow$ & LPIPS$\downarrow$ & PSNR$\uparrow$ & SSIM$\uparrow$ & LPIPS$\downarrow$ \\
			\midrule
			CLDefocus & \textbf{32.16} & \textbf{0.865} & \textbf{0.201} & \textbf{26.62} & \textbf{0.845} & \underline{0.242} & \underline{24.74} & \underline{0.764} & \textbf{0.303} & \underline{24.73} & \underline{0.776} & \underline{0.265} \\
			SYNDOF & 27.58 & 0.790 & 0.296 & 24.55 & 0.743 & 0.261 & 21.64 & 0.654 & 0.474 & 23.77 & 0.743 & 0.315 \\
			DPDD & \underline{28.27} & \underline{0.825} & \underline{0.263} & \underline{25.95} & \underline{0.833} & \textbf{0.207} & \textbf{25.03} & \textbf{0.771} & \underline{0.335} & \textbf{26.11} & \textbf{0.817} & \textbf{0.223} \\
			\bottomrule
	\end{tabular}}
\end{table}
\begin{table}[t]
	\centering
	\setlength{\tabcolsep}{1pt}
	\caption{Quantitative results on no-reference metrics (NIQE~\cite{mittalMakingCompletelyBlind2013}, MUSIQ~\cite{keMUSIQMultiScaleImage2021}, TOPIQ~\cite{chenTOPIQTopDownApproach2024}). The best and second-best results are highlighted in bold and underlined, respectively.}
	\label{tab:result_main_noref}
	\resizebox{1\linewidth}{!}{
		\begin{tabular}{lcccccccccccc}
			\toprule
			Test set & \multicolumn{3}{c}{CLDefocus} & \multicolumn{3}{c}{RTF} & \multicolumn{3}{c}{RealDOF} & \multicolumn{3}{c}{DPDD} \\
			\cmidrule(lr){2-4}\cmidrule(lr){5-7}\cmidrule(lr){8-10}\cmidrule(lr){11-13}
			Train set & NIQE$\downarrow$ & MUSIQ$\uparrow$ & TOPIQ$\uparrow$ & NIQE$\downarrow$ & MUSIQ$\uparrow$ & TOPIQ$\uparrow$ & NIQE$\downarrow$ & MUSIQ$\uparrow$ & TOPIQ$\uparrow$ & NIQE$\downarrow$ & MUSIQ$\uparrow$ & TOPIQ$\uparrow$ \\
			\midrule
			CLDefocus & \underline{7.724} & \textbf{39.850} & \textbf{0.348} & 4.263 & \textbf{59.144} & \underline{0.533} & \textbf{5.029} & \textbf{35.787} & \textbf{0.303} & \textbf{4.657} & \underline{56.696} & 0.487 \\
			SYNDOF & 12.048 & 36.904 & 0.316 & \textbf{3.855} & 57.279 & \textbf{0.539} & 6.285 & 23.905 & 0.241 & 5.067 & 55.819 & \underline{0.493} \\
			DPDD & \textbf{7.508} & \underline{38.961} & \underline{0.345} & \underline{4.196} & \underline{58.822} & 0.531 & \underline{6.030} & \underline{31.350} & \underline{0.270} & \underline{4.746} & \textbf{59.473} & \textbf{0.507} \\
			\bottomrule
	\end{tabular}}
\end{table}

\begin{figure*}[t]
  \centering
  \includegraphics[width=\linewidth]{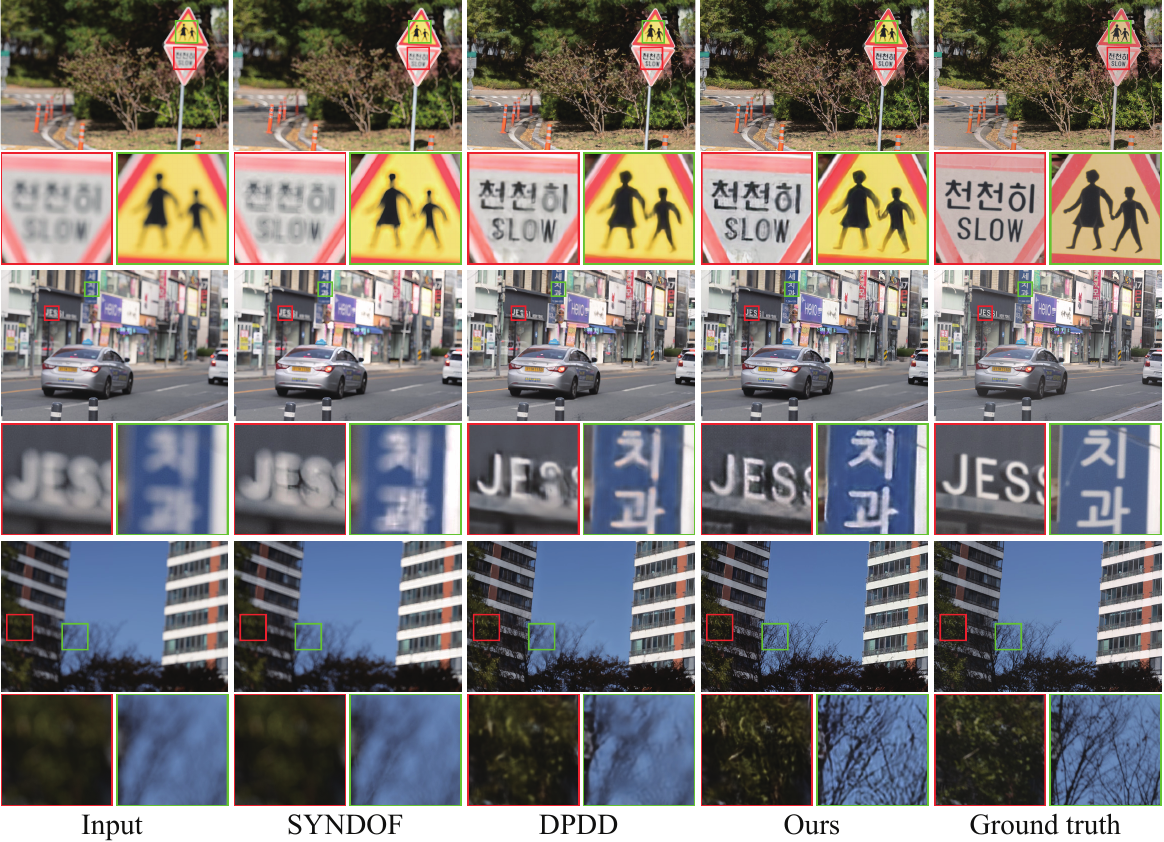}
  \caption{Qualitative comparisons on the RealDOF dataset.}
  \label{fig:qualitative}
\end{figure*}

For comparison, we train NRKNet on three datasets: DPDD~\cite{abuolaimDefocusDeblurringUsing2020}, SYNDOF~\cite{leeDeepDefocusMap2019}, and our CLDefocus dataset.
DPDD is a real-captured dataset collected with a limited camera configuration, whereas SYNDOF is a synthetic dataset constructed using simplified blur models.
For fair comparison, we keep the training protocol identical across datasets.
Each model is evaluated on our CLDefocus test set and existing benchmark datasets: RTF~\cite{dandresNonParametricBlurMap2016}, RealDOF~\cite{leeIterativeFilterAdaptive2021}, and DPDD.

\cref{tab:result_main_fullref} presents the full-reference results, while \cref{tab:result_main_noref} reports the no-reference results.
On our test set, the CLDefocus-trained model achieves the best performance under both metric types, confirming the physical consistency and diversity of our dataset.
On real benchmarks, the CLDefocus-trained model generally yields lower full-reference scores than the DPDD-trained model.
This behavior is attributable to imperfections in real ground-truth images, including residual defocus, brightness inconsistencies, and spatial misalignment, which bias pixel-wise measures such as PSNR and SSIM.
We analyze the impact of these ground-truth imperfections in \cref{subsec:limitation_gt}.

In contrast, the CLDefocus-trained model achieves overall higher no-reference scores, implying improved perceptual quality.
This trend is also observed in \cref{fig:qualitative}, where our model restores sharper details with fewer artifacts.
The DPDD-trained model tends to overfit to its specific lens configuration, limiting cross-lens generalization, while the SYNDOF-trained model generalizes poorly due to its simplified blur model.
Overall, CLDefocus enhances cross-device generalization through physically grounded PSF modeling and scalable lens diversity.
Additional results in the supplementary material further support this trend.

\subsection{Limitations of Ground-Truth in Real Datasets}
\label{subsec:limitation_gt}

\begin{figure*}[t]
  \centering
  \includegraphics[width=\linewidth]{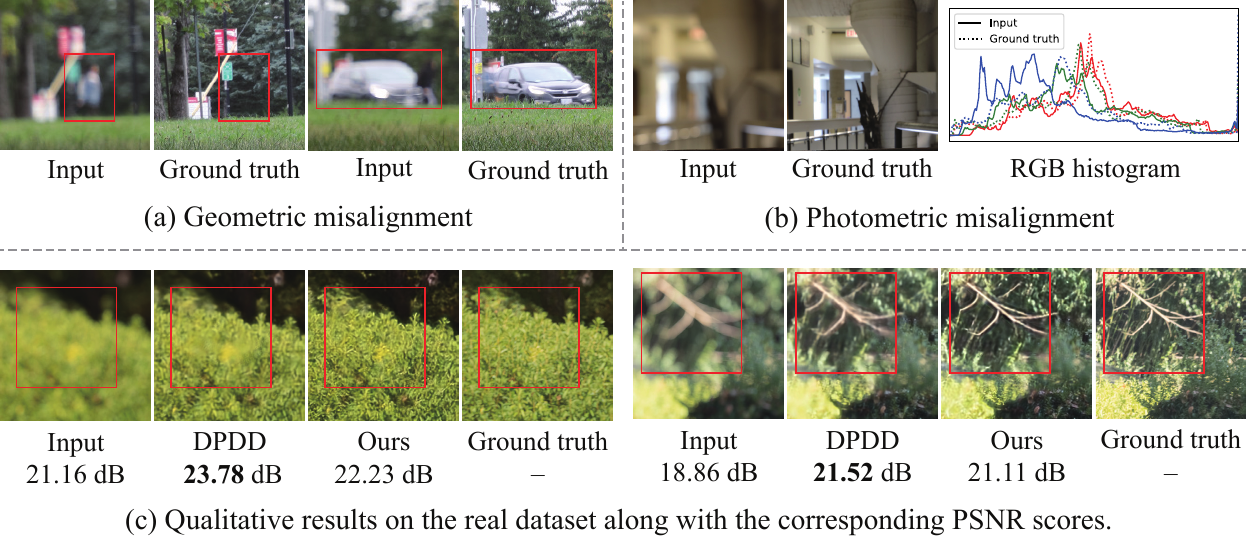}
  \caption{Limitations of ground truth in real-captured defocus datasets.
  Geometric misalignment caused by dynamic objects and photometric misalignment introduced by aperture changes lead to spatial and color inconsistencies, as reflected in the RGB histograms.
  These imperfections result in a mismatch between full-reference metrics and visual quality, where perceptually sharper restorations may receive lower metric scores.}
  \label{fig:real_gt_limitation}
\end{figure*}

We analyze structural imperfections in real-captured ground-truth pairs.
Real-captured defocus datasets inevitably contain residual inconsistencies between input and ground-truth (GT) images. 
\cref{fig:real_gt_limitation}-(a) illustrates geometric misalignment: scene objects may shift, deform, or newly appear between captures due to subject motion or scene dynamics. 
Even in carefully captured datasets such as DPDD~\cite{abuolaimDefocusDeblurringUsing2020} or post-processed datasets such as RealDOF~\cite{leeIterativeFilterAdaptive2021}, such spatial inconsistencies remain. 
\cref{fig:real_gt_limitation}-(b) further shows photometric misalignment. Changing apertures between captures often introduces brightness and color variations~\cite{leeIterativeFilterAdaptive2021}. 
These discrepancies are visually evident and are reflected in the RGB histograms, where the channel-wise intensity distributions differ significantly between input and GT images. 
Moreover, some GT images still contain residual defocus blur, indicating that the GT itself may not represent a perfectly focused reference.

These inconsistencies directly affect full-reference evaluation. 
As shown in \cref{fig:real_gt_limitation}-(c), even when our result is perceptually sharper, pixel-wise metrics such as PSNR can favor predictions that remain slightly blurred, as they better match misaligned or partially blurred GT images. 
Consequently, models trained on such datasets may implicitly learn a conservative, blur-preserving bias, yielding higher full-reference scores despite inferior perceptual sharpness.

\subsection{Debye CZT vs. Huygens' Principle}

\begin{figure*}[t]
  \centering
  \includegraphics[width=\linewidth]{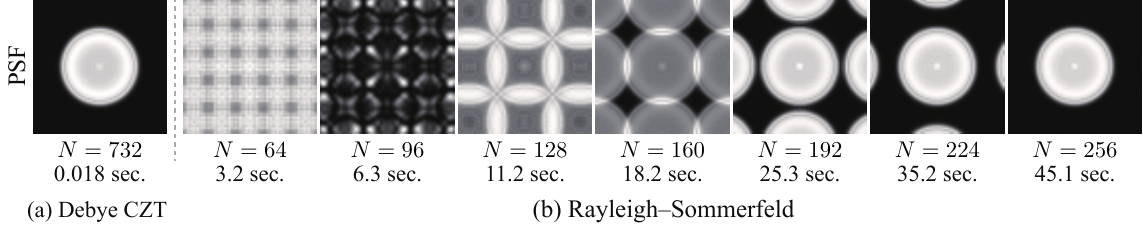}
  \caption{Comparison between the Debye CZT and the Rayleigh--Sommerfeld. The Debye CZT determines the sampling density and produces a stable, aliasing-free PSF without empirical tuning. In contrast, the Rayleigh--Sommerfeld provides no explicit sampling criterion, so we adjust the sampling density manually. Insufficient sampling leads to severe PSF warping and geometric distortion. Only sufficiently dense sampling yields a valid PSF, but at the cost of increased runtime, as indicated for each configuration.}
  \label{fig:sampling_condition}
\end{figure*}

We examine the stability and computational efficiency of our Debye CZT framework by comparing it with a propagation method based on the Huygens' principle, which has been adopted in prior wave-optics PSF simulations~\cite{chenOpticalAberrationsCorrection2021, luoCorrectingOpticalAberration2024}.
For this, we implement a method using the Rayleigh--Sommerfeld integral~\cite{marathayUsualApproximationUsed2004}.

\subsubsection{Sampling Condition.}

To validate the stability of our Debye CZT, we compare it with the Rayleigh--Sommerfeld with various sampling numbers.
In both cases, we first construct the Zernike wavefront on the exit pupil sphere from the same lens design.
The difference lies in the propagation step to the sensor plane: our method performs propagation via the Debye CZT under the explicit sampling condition derived in \cref{subsec:debye}, whereas the Rayleigh--Sommerfeld directly sums the contribution of each sampled wavefront point to every sensor-plane point.
We use ``JP2017-003807\_Example04P.zmx'' in the lens collection~\cite{lensdesigns}.
The PSF is computed for an on-axis point at \qty{0.8}{\meter}, with the sensor focused at infinity. We set the spatial extent of a PSF to $64 \times 64$.
For a fair comparison, we use $5\times$ supersampling for the PSF computation in both methods.

For the Debye CZT, \cref{eqn:debye_ft_sampling} yields $\DebyeSamBound = 366$, and we set $\InSam = 2\DebyeSamBound = 732$ following our strategy described in \cref{subsec:debye}.
In contrast, as the Rayleigh--Sommerfeld does not provide well-known explicit sampling criteria, we start from $\InSam = 64$ and gradually increase it by $32$ to reach $256$.
As shown in \cref{fig:sampling_condition}, only $\InSam = 256$ produces a valid PSF, while smaller $\InSam$ values exhibit aliasing manifested as PSF warping.
This implies that we also need a proper sampling condition like \cref{eqn:debye_ft_sampling} for the Rayleigh--Sommerfeld.

\subsubsection{Computational Cost.}

We measured the runtime of PSF computation after evaluating wavefront samples. 
The Debye CZT took \qty{0.018}{\second}, whereas the Rayleigh--Sommerfeld took \qty{45.1}{\second} for $\InSam = 256$.
Computation times for other $\InSam$ values are stated in \cref{fig:sampling_condition}.
The theoretical complexity of the Debye CZT is $\mathcal{O}((\InSam + \OutSam)^2 \log(\InSam + \OutSam))$, while the Rayleigh--Sommerfeld scales as $\mathcal{O}(\InSam^2 \OutSam^2)$,
where $\OutSam$ is the output grid size.
Its quadratic dependence on both $\InSam$ and $\OutSam$ makes the Rayleigh--Sommerfeld highly sensitive to sampling density.
Detailed derivations and complexity analysis are provided in the supplementary material.

\subsection{Examples of Generated PSFs}

\begin{figure*}[t]
  \centering
  \includegraphics[width=\linewidth]{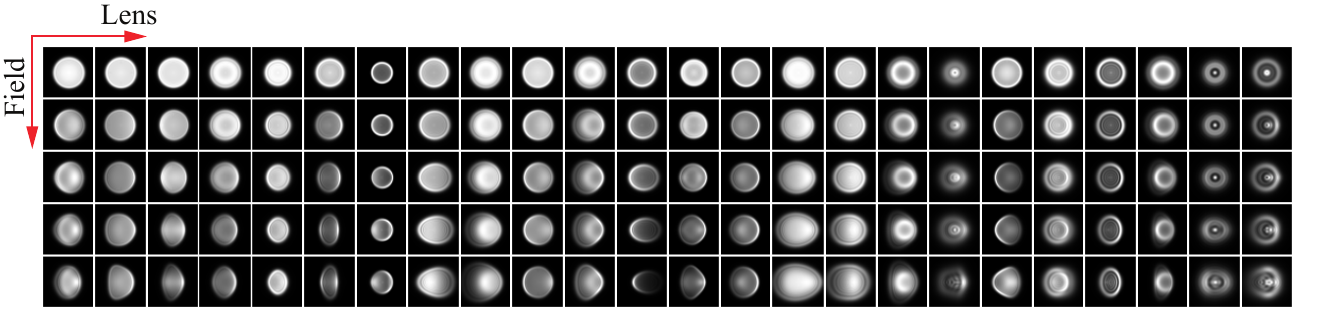}
  \caption{
  PSF examples of various lens designs (column) and field positions (row), computed by the Debye CZT.
  }
  \label{fig:psf_examples}
\end{figure*}

\cref{fig:psf_examples} shows examples of diverse PSFs computed by the Debye CZT.
They show diverse shapes and textures of PSFs across different settings.
In particular, PSFs of severe off-axis fields often have complex and asymmetric shapes.
These imply that simplified blur models are insufficient to simulate realistic defocus blur.
We provide more examples and analysis in the supplementary material.

\subsection{Ablation Study}

We conduct an ablation study to evaluate the contributions of three components in our synthesis pipeline: (1) realistic lens PSFs, (2) the imaging pipeline, and (3) depth-varying scenes.
For each experiment, we maintain all other components, including consistent patch sampling and augmentations, as much as possible.

\subsubsection{Realistic Lens PSF.}

We first investigate the impact of the lens PSFs by replacing them with isotropic Gaussian blur.
For fair comparison, we match the blur magnitude of the Gaussian kernels to the original PSFs by deriving their standard deviation $\sigma$ from the circle of confusion (CoC).
Following prior works~\cite{leeDeepDefocusMap2019, nazirDepthEstimationImage2023}, we set $\sigma = \textrm{CoC radius} / 2$.
As shown in \cref{tab:ablation}, replacing lens PSFs with Gaussian kernels leads to a significant degradation in performance.
Qualitatively, \cref{fig:ablation} reveals that substantial blur remains in the restored images, likely because the model trained with Gaussian blur fails to generalize to complex real blur.

\subsubsection{Realistic Imaging Pipeline.}

Next, we analyze the impact of the realistic imaging pipeline.
We remove the ISP and noise synthesis in the RAW space.
In this setup, we directly sample a sharp RGB patch, synthesize blur in RGB space, and obtain the blurred RGB patch without ISP.
The results in \cref{tab:ablation} demonstrate that the absence of a realistic imaging pipeline results in a notable performance drop, highlighting its importance.

\subsubsection{Depth-Varying Scenes.}

\begin{figure*}[t]
  \centering
  \includegraphics[width=\linewidth]{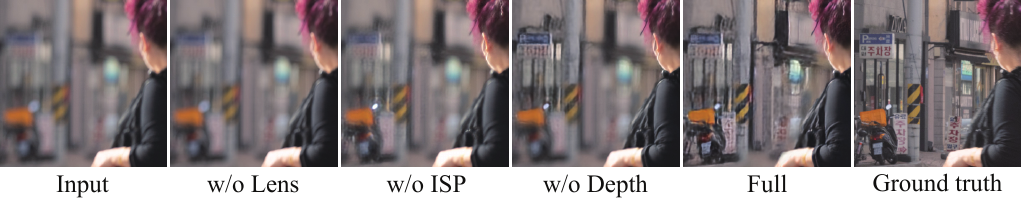}
  \caption{Qualitative ablation results illustrating the impact of each component.}
  \label{fig:ablation}
\end{figure*}

Finally, we examine the impact of depth-varying scenes.
We simplify each patch into a single layer, allowing for blur synthesis via a uniform convolution.
As shown in \cref{tab:ablation}, this leads to only a marginal decrease in overall metrics, while the model remains vulnerable to defocus discontinuities.
Specifically, as shown in \cref{fig:ablation}, the model exhibits concentrated artifacts near edges with sharp defocus transitions.
This suggests that modeling depth variance is crucial to handle discontinuous defocus changes near object boundaries.
\begin{table}[t]
\centering
\setlength{\tabcolsep}{2pt}
\caption{Ablation results on the RealDOF dataset showing the impact of each component. The \colorbox{red!30}{best}, \colorbox{orange!25}{second best}, and \colorbox{yellow!25}{third best} results are highlighted.}
\label{tab:ablation}
\begin{tabular}{lcccccc}
\toprule
Components & PSNR$\uparrow$ & SSIM$\uparrow$ & LPIPS$\downarrow$ & NIQE$\downarrow$ & MUSIQ$\uparrow$ & TOPIQ$\uparrow$ \\
\midrule
	w/o Lens & 22.34 & 0.667 & 0.521 & 7.733 & 25.162 & 0.238 \\
	w/o ISP & \cellcolor{yellow!25}23.49 & \cellcolor{yellow!25}0.746 & \cellcolor{yellow!25}0.349 & \cellcolor{orange!25}5.191 & \cellcolor{yellow!25}32.045 & \cellcolor{yellow!25}0.285 \\
	w/o Depth & \cellcolor{orange!25}24.54 & \cellcolor{orange!25}0.762 & \cellcolor{orange!25}0.317 & \cellcolor{yellow!25}5.292 & \cellcolor{orange!25}34.999 & \cellcolor{orange!25}0.300 \\
	Full & \cellcolor{red!30}24.74 & \cellcolor{red!30}0.764 & \cellcolor{red!30}0.303 & \cellcolor{red!30}5.029 & \cellcolor{red!30}35.787 & \cellcolor{red!30}0.303 \\
\bottomrule
\end{tabular}
\end{table}

\section{Conclusion}

In this paper, we proposed a framework for synthesizing realistic defocus deblurring datasets for compound lenses.
It integrates efficient wave-optics PSF computation, depth-aware defocus rendering with occlusion handling, and ISP-aware blur synthesis in the radiometrically linear space.
This unified framework enables scalable, photorealistic dataset generation with improved computational efficiency over existing approaches.
We also presented the CLDefocus dataset and showed that it is competitive with the real-captured DPDD for training.

\paragraph{Limitations and Future Work.}
Our pipeline still has limitations.
The Debye CZT may yield less accurate PSFs for severe off-axis fields due to the Debye approximation and failure to capture complex pupil shapes.
Moreover, imprecise depth maps might produce implausible synthesis results, and the simple ISP model might not fully capture the diverse ISPs of real-world cameras.
The pipeline offers flexibility that remains largely unexplored. 
Its controllable parameters enable task-specific dataset design. 
For example, blur levels can be adjusted to construct datasets for curriculum learning, which is difficult to achieve with captured data. 
We hope future work will further leverage this flexibility for broader tasks.

\section*{Acknowledgments}
This work was supported by the National Research Foundation of Korea (NRF) (RS-2026-25492695; Basic Science Research Program, RS-2022-NR070870) and by the Institute of Information \& Communications Technology Planning \& Evaluation (IITP) (RS-2019-II191906, Artificial Intelligence Graduate School Program (POSTECH); IITP-2026-RS-2024-00437866, ITRC Program), funded by the Korea government (MSIT).

\clearpage
\title{Realistic Compound-Lens Defocus Blur Synthesis \textnormal{Supplementary Material}}
\titlerunning{Realistic Compound-Lens Defocus Blur Synthesis}
\author{Yunkyu Lee\inst{1}\orcidlink{0009-0005-5375-9330} \and
Woohyeok Kim\inst{1}\orcidlink{0009-0006-5691-5447} \and
Sunghyun Cho\inst{1}\orcidlink{0000-0001-7627-3513}}
\authorrunning{Y.~Lee et al.}
\institute{POSTECH, Pohang, Korea\\
\email{\{lyk1012,woohyeok,s.cho\}@postech.ac.kr}}
\maketitle

\setcounter{section}{0}
\setcounter{figure}{0}
\setcounter{table}{0}
\setcounter{equation}{0}
\renewcommand{\thesection}{S\arabic{section}}
\renewcommand{\thesubsection}{S\arabic{section}.\arabic{subsection}}
\renewcommand{\thetable}{S\arabic{table}}
\renewcommand{\thefigure}{S\arabic{figure}}
\renewcommand{\theequation}{S\arabic{equation}}
\renewcommand{\theHsection}{S\arabic{section}}
\renewcommand{\theHsubsection}{S\arabic{section}.\arabic{subsection}}
\renewcommand{\theHtable}{S\arabic{table}}
\renewcommand{\theHfigure}{S\arabic{figure}}
\renewcommand{\theHequation}{S\arabic{equation}}

\setcounter{section}{0}
\setcounter{figure}{0}
\setcounter{table}{0}

\noindent
This supplementary material provides additional details and results that complement the main paper.
It includes
(1) detailed descriptions of our method omitted from the main paper,
(2) additional experimental results.

\begin{itemize}
    \item \textbf{PSF formulation and wave-optics modeling (\cref{sec:supp_lens_psf}).}
    We concretely define the PSF computation problem and describe the general wave-optics formulation for compound lenses.
    
    \item \textbf{Derivation of Debye CZT and comparisons (\cref{sec:supp_debye}).}
    We present the derivation and implementation details of the Debye CZT and compare it with two alternative PSF computation methods.
    
    \item \textbf{Lens filtering and diversity (\cref{sec:supp_lens_designs}).}
    We describe the criteria used to filter compound lens designs when constructing the lens library, which are omitted from the main paper.
    Moreover, we provide a brief analysis of the lens diversity.
    
    \item \textbf{ISP-aware defocus synthesis (\cref{sec:supp_synthesis}).}
    We detail the image signal processor (ISP) model used in our pipeline and explain the synthesis processes for noise and saturation.
    
    \item \textbf{Extended experimental results (\cref{sec:supp_other_deblurring}).}
    We provide additional experimental results that extend the evaluation in the main paper to more deblurring models, evaluation metrics, and qualitative examples.
    
    \item \textbf{Evaluation on a smartphone (\cref{sec:supp_smartphone}).}
    We provide additional evaluation results on images captured by a smartphone camera.
    
    \item \textbf{Downstream vision tasks (\cref{sec:supp_downstream}).}
    We provide experimental results on downstream vision tasks.
\end{itemize}

\section{Computing PSFs of Compound Lenses}
\label{sec:supp_lens_psf}

\providecommand{\RayOfT}{\vec{p}_t}
\providecommand{\ImplSurf}{S}
\providecommand{\AngleInc}{\theta_i}
\providecommand{\AngleTrans}{\theta_t}
\providecommand{\IORInc}{n_i}
\providecommand{\IORTrans}{n_t}
\providecommand{\Normal}{\vec{n}}

\providecommand{\RefSphere}{S}
\providecommand{\RefPt}{F}
\providecommand{\Imag}{j}

\providecommand{\Upsam}{u}
\providecommand{\KerSize}{k}

\subsection{Problem Definition}
\label{subsec:supp_psf_problem}

Here, we provide a more concrete definition of the problem than in the main paper.
Our goal is to generate the point spread function (PSF), defined as the image formed on the sensor by a point source.
The light emitted from the source is collected by the optical system to form an image.
A sensor then samples this image on a discrete grid.
The focusing distance determines the sensor position that brings a source at that distance into focus.
Defocus occurs when the source depth differs from the focusing distance.

In practical photography, the effective range of the PSF is much smaller than the whole sensor.
Hence, it is efficient to view only a small $\KerSize \times \KerSize$ viewport.
We refer to a viewport as a small subgrid on the sensor grid that contains all significant parts of the PSF.
We also consider the implications of sampling theory.
If we take only one sample to determine each grid pixel, we may encounter aliasing.
Since diffraction patterns tend to change rapidly, this may significantly reduce quality.
To address this, we simply upsample the viewport by a factor $\Upsam > 1$ and downsample it later.
In our experiments, we used $\Upsam = 5$, which is sufficient to prevent aliasing across almost all generation results.

\subsection{Ray Tracing}
\begin{figure}[t]
  \centering
  \includegraphics[width=\linewidth]{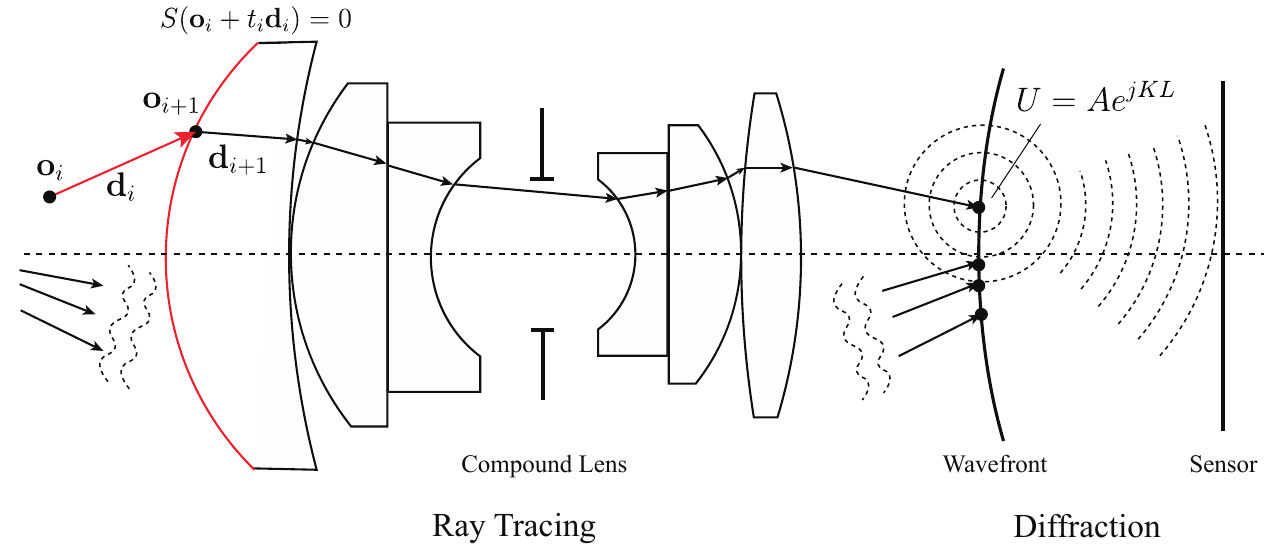}
  \caption{
  An illustration of the common process for simulating wave-based imaging of a compound lens.
  Inside the lens, light propagation is simulated via ray tracing.
  After passing through the last lens surface, the rays are projected onto a reference surface and converted into a wavefront.
  Under wave optics, a diffraction model propagates the wavefront to the sensor plane, yielding the final image.
  }
  \label{fig:supp_lens}
\end{figure}

A ray is a convenient abstraction for analyzing light propagation within optical systems.
Ray tracing through compound lenses has been employed in many prior works related to optics~\cite{chenOpticalAberrationsCorrection2021, tsengDifferentiableCompoundOptics2021, wangDifferentiableEngineDeep2022, yangCurriculumLearningInitio2024}.
It can be regarded as an iteration of two steps: (1) propagation to the next surface and (2) refraction at the intersection with that surface, as illustrated in \cref{fig:supp_lens}.

For ray propagation, consider a ray defined as $ \RayOfT = \vec{o} + t \vec{d} $ originating from $\vec{o} \in \mathbb{R}^3$ with the direction $\vec{d} \in \mathbb{R}^3$.
Assume that the next surface is defined by $\ImplSurf(\vec{p}) = 0$ using an implicit surface function $\ImplSurf: \mathbb{R}^3 \to \mathbb{R}$.
The propagation step seeks to find $t \ge 0$ such that:
\begin{equation}
    \ImplSurf(\RayOfT) = \ImplSurf(\vec{o} + t \vec{d}) = 0 ,
\end{equation}
which can be numerically solved using iterative algorithms such as Newton's method for general surfaces.
The next ray origin is determined as \( \vec{o}' = \vec{o} + t \vec{d} \).

After propagating the ray to the intersection point, we perform the refraction step.
Snell's law:
\begin{equation}\label{eqn:supp_snell}
    \IORTrans \sin \AngleTrans = \IORInc \sin \AngleInc
\end{equation}
determines the refracted angle $\AngleTrans$ from the incident angle $\AngleInc$, where $\IORInc$ and $\IORTrans$ denote the indices of refraction for the incident and transmission media, respectively, and $\Normal$ denotes the normal vector at the intersection point.
Note that $\Normal$ can be calculated with the gradient of $\ImplSurf$.
To obtain the direction vector $\vec{d}'$ from $\AngleTrans$, we use the formula~\cite{chenOpticalAberrationsCorrection2021}:
\begin{equation}\label{eqn:supp_refraction_direction}
    \vec{d}' = \eta \, \vec{d} + ( \eta \cos \AngleInc - \cos \AngleTrans ) \, \Normal ,
\end{equation}
where \( \eta = \IORInc / \IORTrans \).
The solution of \cref{eqn:supp_refraction_direction} may be undefined, which is physically interpreted as total internal reflection.
In such cases, we disregard these rays.

We implemented a ray tracer specifically for spherical and even-polynomial aspheric surfaces, which cover most lenses in the lens database that we used.
We excluded design files that use other surface types.

\subsection{Wave-Optics Simulation of Compound Lenses}
\label{subsec:supp_wave_optics_of_lens}

For wave-based imaging of a compound lens, it is common practice to reconstruct a wavefront from traced rays and subsequently apply diffraction models.
To achieve this, we first establish a reference surface to define the wavefront, as illustrated in \cref{fig:supp_lens}.
To reconstruct wavefront samples, we also calculate the optical path length (OPL) of each ray during the ray-tracing process.
The OPL is initialized at $0$ at the point source and accumulated as $n t$ for each propagation step, where $n$ is the index of refraction and $t$ is the propagation distance.
After tracing rays toward the last lens surface, we finally trace them to the designated reference surface.
A ray with OPL $L$ is converted to a wavefront sample $A \exp(\Imag k L)$, where $\Imag = \sqrt{-1}$ and $k$ denotes the wave number.
$A$ denotes the amplitude, which we assume to be constant for simplicity.
Using the reconstructed wavefront, we can employ diffraction models such as the Rayleigh--Sommerfeld integral or the Debye integral (used in our method) to compute the light wave on the image plane.

This approach is frequently referred to as the ``hybrid method'' and has been adopted by most prior works dealing with the wave-optics imaging of compound lenses~\cite{chenOpticalAberrationsCorrection2021, luoCorrectingOpticalAberration2024, hoDifferentiableWaveOptics2025, mullerExaminingImpactOptical2026}.
Zemax's ``Huygens PSF'', which is often used as an accurate reference, also relies on this principle.
This method ultimately assumes that diffraction occurs only once at the exit pupil boundary.
While this introduces a fundamental error by neglecting internal diffractions, it is considered the most accurate practical approach for computing the PSFs of compound lenses in many applications.

As stated in \cref{subsec:supp_psf_problem}, our objective is to acquire wave intensity values on a viewport grid.
Our method (Debye CZT) allows for the effective output-sensitive evaluation on that grid.

\section{Detailed Description of Debye CZT}
\label{sec:supp_debye}

\providecommand{\NA}{\mathrm{NA}}
\providecommand{\XPCen}{P}
\providecommand{\XPRad}{R}
\providecommand{\FocPt}{F}
\providecommand{\FocLen}{f}
\providecommand{\ImgZ}{z_\mathrm{img}}
\providecommand{\InWave}{\mathbf{E}_t}
\providecommand{\OutWave}{\mathbf{E}}
\providecommand{\RefSphr}{S}
\providecommand{\CapDomain}{\Omega}
\providecommand{\OffAxisAngle}{\alpha}
\providecommand{\Wvl}{\lambda}

\subsection{Debye Diffraction Formulation}

\begin{figure}[t]
  \centering
  \includegraphics[width=\linewidth]{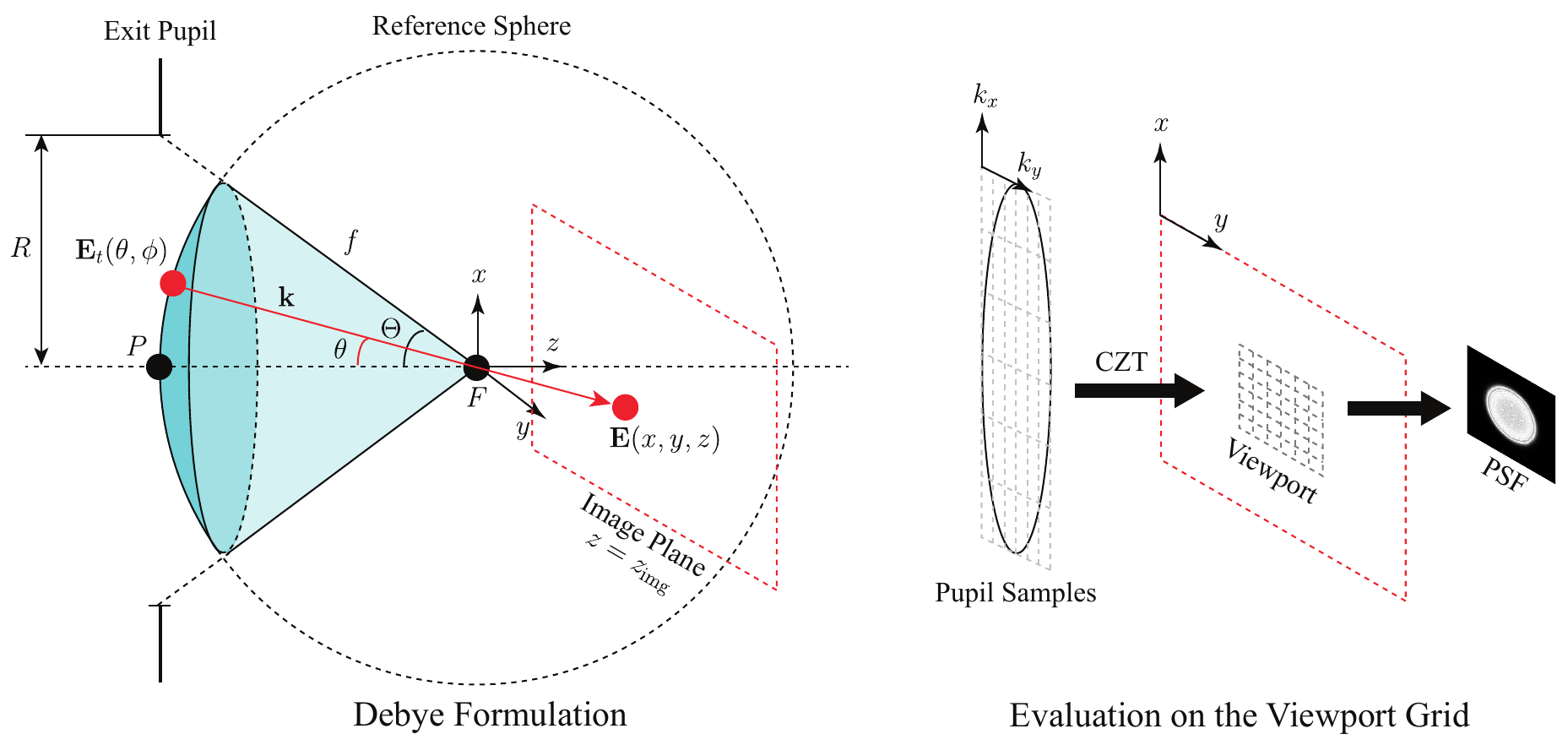}
  \caption{
  A figure illustrating the Debye CZT in detail.
  The left side shows the physical setup used in the Debye formulation.
  The reference sphere $\RefSphr$ is centered at the focal point $\FocPt$ and passes through the exit pupil center $\XPCen$.
  Its radius equals the focal length $\FocLen = \overline{\XPCen \FocPt}$.
  The focal point $\FocPt$ is set as the origin.
  $\mathbf{k}$ denotes the wave vector corresponding to $(\theta, \phi)$.
  The shaded region represents the spherical cap with angle $\Theta = \arctan(\XPRad / \FocLen)$, where $\XPRad$ is the exit pupil radius.
  The right side illustrates the CZT that converts pupil samples into the output wave observation on the viewport grid.
  }
  \label{fig:supp_debye}
\end{figure}

There are several diffraction models that deal with different reference surfaces.
Many models, including Fresnel propagation, the angular spectrum method, and the Rayleigh--Sommerfeld integral (for planar surfaces), typically use the exit pupil plane as the reference surface.
However, we find these are not suitable for efficiently rendering PSFs with large defocus.
Photographic lenses aim for precise imaging, meaning their output waves approximately form a converging spherical wave toward a focal point.
If these waves are observed on a planar surface, they exhibit rapid quadratic phase variations.
This necessitates a very high sampling density to avoid aliasing artifacts.

An effective alternative is to use a spherical reference surface aligned with the converging output wave.
This ensures that the observed wavefront is significantly simpler.
In an ideal imaging system, where the wavefront is a perfect converging spherical wave, choosing a matching spherical reference surface results in constant phase.
In real optical systems, aberrations introduce phase variations; however, these vary much more slowly than the quadratic phases encountered with planar reference surfaces.

The Debye formulation~\cite{leuteneggerFastFocusField2006} is a diffraction model designed for such spherical references.
The physical setup is illustrated on the left side of \cref{fig:supp_debye}.
It models the diffraction of a wavefront on a \emph{cap} of an on-axis sphere with a half-angle $\Theta$.
In this model, the input wave $\InWave(\theta, \phi)$ on the cap behaves as a plane wave spectrum with the wave vector $\mathbf{k}$.
The resulting output wave $\OutWave(x, y, z)$ is defined as~\cite{leuteneggerFastFocusField2006}:
\begin{equation}\label{eqn:supp_debye_integral}
\OutWave(x, y, z) = -\frac{\Imag \FocLen}{\Wvl} \iint_\CapDomain \InWave(\theta, \phi) \exp(\Imag \mathbf{k} \cdot (x, y, z)) \, \dd \CapDomain ,
\end{equation}
where $\CapDomain$ denotes the domain of the spherical cap.
Following the notation of \cite{leuteneggerFastFocusField2006}, $\mathbf{k}$ is defined as:
\begin{equation}\label{eqn:supp_wave_vector}
\mathbf{k} = (-k_x, -k_y, k_z) = k (-\cos\phi \sin\theta, -\sin\phi \sin\theta, \cos\theta) ,
\end{equation}
where $k = k_0 n_t$, $k_0 = 2\pi / \Wvl$ is the wave number, and $n_t$ is the refractive index of the medium (set to $1$ in our case).
While \cref{eqn:supp_debye_integral} allows for the evaluation of $\OutWave$ at an arbitrary $(x, y, z)$, in practice, we evaluate it on an image plane at $z = \ImgZ$ near the focal point $\FocPt$.

To compute lens PSFs using the Debye formulation, we define the reference sphere as the exit pupil sphere.
This sphere is centered at the on-axis focal point $\FocPt$ and passes through the center of the exit pupil $\XPCen$.
As shown in \cref{fig:supp_debye}, the coordinates are shifted to place $\FocPt$ at the origin for convenience.
The focal length $\FocLen$ is the distance between $\XPCen$ and $\FocPt$, which corresponds to the radius of the reference sphere.
The cap angle, which determines the numerical aperture (NA), is defined as $\Theta = \arctan(\XPRad / \FocLen)$, where $\XPRad$ is the exit pupil radius.
As explained in \cref{subsec:supp_wave_optics_of_lens}, $\InWave(\theta, \phi)$ is determined by the rays traced to the reference sphere at coordinates $(\theta, \phi)$.
A key consideration is the definition of the focal point $\FocPt$.
In an aberration-free lens, all rays converge to a single point, which naturally becomes $\FocPt$.
In general, however, rays do not intersect at a single point due to aberrations.
Therefore, we define $\FocPt$ as the point of best intersection for the rays, obtained by solving the least-squares problem:
\begin{equation}
    \vec{p}^\ast = \argmin_{\vec{p} \in \mathbb{R}^3} \sum_{i=1}^N w_i \, \mathrm{dist} (\vec{p}, r_i)^2 ,
\end{equation}
where $r_1, \ldots, r_N$ are the traced rays with associated weights $w_1, \ldots, w_N$, and $\mathrm{dist}$ denotes the Euclidean distance between a point and a ray (line).
While other weighting schemes based on radiance are possible, we simply set $w_i = 1$ for all valid rays.

\subsection{Computing Debye Diffraction via CZT}
\label{subsec:supp_debye_czt}

\subsubsection{Derivation of the Fourier Form.}

A straightforward approach to computing \cref{eqn:supp_debye_integral} is direct numerical integration.
However, as noted by \cite{leuteneggerFastFocusField2006}, this approach incurs a high computational cost.
Alternatively, \cite{leuteneggerFastFocusField2006} reformulated \cref{eqn:supp_debye_integral} into a Fourier transform, enabling more efficient numerical methods.
We briefly summarize the derivation of this Fourier form.
\cite{leuteneggerFastFocusField2006} reparameterized the integration variables of \cref{eqn:supp_debye_integral} into $(k_x, k_y)$, using the relation:
\begin{equation}
    \dd \CapDomain = \frac{1}{k^2} \frac{\dd k_x \dd k_y}{\cos \theta} .
\end{equation}
This yields an integral in the form of the Fourier transform with respect to $k_x$ and $k_y$:
\begin{equation}
    \OutWave(x, y, z) = -\frac{\Imag \FocLen}{\Wvl k^2} \iint\limits_{k_{xy} < k \sin\Theta} \frac{\InWave(\theta, \phi) \exp(\Imag k_z z)}{\cos\theta} \exp(-\Imag (k_x x + k_y y)) \, \dd k_x \dd k_y ,
\end{equation}
where $k_{xy} = \sqrt{k_x^2 + k_y^2}$.
This can be directly expressed in Fourier form as:
\begin{equation}\label{eqn:supp_debye_ft}
    \OutWave(x, y, z) = -\frac{\Imag \FocLen}{\Wvl k^2} \mathcal{F} \left[ \frac{\InWave(\theta, \phi) \exp(\Imag k_z z)}{\cos\theta} \right] .
\end{equation}
Note that $k_z$, $\theta$, and $\phi$ are determined by $k_x$ and $k_y$ via \cref{eqn:supp_wave_vector}.
For a detailed analysis, please refer to \cite{leuteneggerFastFocusField2006}.
Although the formulation is originally based on vector diffraction theory, we reduce it to scalar diffraction by replacing vector waves with scalar waves.
This simplification is justified because photographic lenses typically have a small NA, meaning vectorial effects are negligible.

\subsubsection{The Chirp Z-transform (CZT).}

While \cref{eqn:supp_debye_ft} suggests numerical methods based on the Fourier transform, standard algorithms such as the fast Fourier transform (FFT) are not ideally suited for our purposes.
As explained in \cref{subsec:supp_psf_problem}, we need to observe the image wave on a small viewport with an arbitrary offset and a specific pixel pitch (sensor-aligned grid).
However, the FFT imposes strict constraints on the output grid.
Consider a one-dimensional FFT with an input grid of size $N$ and sample pitch $\delta_x$ centered at the origin.
The resulting output grid is fixed to a size $N$ with a sample pitch $\delta_f = 1 / (N \delta_x)$, and is also centered at the origin~\cite{schmidtNumericalSimulationOptical2010}.
These restrictions limit the utility of the FFT for our arbitrary viewport requirements.

Fortunately, prior works~\cite{leuteneggerFastFocusField2006, huEfficientFullpathOptical2020} have proposed a solution: the chirp Z-transform (CZT).
The CZT is a generalization of the discrete Fourier transform (DFT) that allows for arbitrary output grid configurations, making it a perfect fit for our goal.
The CZT is defined by the following problem (shown in 1D for simplicity):
\begin{equation}
    Z_n = \sum_{m=0}^{M-1} z_m a^{-m} w^{mn} ,
\end{equation}
where $\{ z_m \}$ is a complex-valued signal and $a$, $w$ are complex-valued constants.
Note that setting $a = 1$ and $w = \exp(-\Imag \Delta k)$ yields the DFT as a special case.
The CZT reformulates this by introducing a chirp factor, which refers to a factor with quadratic phases.
By utilizing the algebraic identity $2mn = n^2 + m^2 - (n-m)^2$, the exponential term can be decomposed as:
\begin{equation}
    Z_n = w^{n^2/2} \sum_{m=0}^{M-1} \left( z_m a^{-m} w^{m^2/2} \right) \cdot w^{-(n-m)^2/2} .
\end{equation}
This can be concisely expressed as a convolution:
\begin{equation}
    Z_n = \left[ \left( z_m a^{-m} w^{m^2/2} \right) \ast \left( w^{-m^2/2} \right) \right] w^{n^2/2} .
\end{equation}
By the convolution theorem, we can compute this efficiently using the FFT.
Specifically, we apply two FFTs to each operand, perform an entrywise product, and then apply an inverse FFT.
Consequently, the operation is completed with two forward FFTs and one inverse FFT.
Because this computation involves a circular convolution, it would result in wrap-around artifacts unless the input grids are zero-padded.
When the input grid is of length $M$ and the desired output grid is of length $N$, the input must be zero-padded to a length of at least $M + N - 1$.

\subsubsection{Computational Cost of the CZT.}

The CZT is computed via two forward FFTs and one inverse FFT.
For 2D cases with $N \times N$ input (pupil) and $M \times M$ output (image) grids, the CZT can be applied separably along the rows and columns.
Its time complexity is $\mathcal{O}((M+N)^2 \log(M+N))$, which is significantly more efficient than direct integration, which takes $\mathcal{O}(N^2 M^2)$.
One drawback of the CZT is that it generally requires 64-bit precision due to the rapidly varying chirp factors accumulating large phase values across the grids.
However, many modern devices support this with GPU acceleration, still allowing for efficient evaluation.

\subsection{Zernike Wavefront Modeling}
\label{subsec:supp_zernike}

\providecommand{\OurZernN}{15}

To compute the Debye formulation via CZT, a uniform wave sampling grid is required on the exit pupil sphere.
However, it is generally difficult to obtain traced rays that coincide exactly with these specific grid points.
To address this, we first construct a continuous wavefront function from the initial ray-traced samples and then resample the wavefront onto the required grid for the CZT.
For this purpose, we use Zernike polynomials~\cite{niuZernikePolynomialsTheir2022}, which are standard basis functions in optical engineering.
For integers $n$ and $m$, a Zernike polynomial is defined as:
\begin{equation}
Z_n^m(\rho,\phi)=
\begin{cases}
R_n^{m}(\rho)\,\cos(m\phi), & m \ge 0,\\[6pt]
R_n^{|m|}(\rho)\,\sin(|m|\phi), & m < 0,
\end{cases}
\end{equation}
on the domain $ 0 \le \rho \le 1, \; 0 \le \phi < 2\pi $, where the radial polynomial $R_n^{\abs{m}}$ is defined as:
\begin{equation}
R_n^{\abs{m}}(\rho)=
\sum_{k=0}^{(n-\abs{m})/2}
(-1)^k
\frac{(n-k)!}
{k!\left(\frac{n+\abs{m}}{2}-k\right)!\left(\frac{n-\abs{m}}{2}-k\right)!}
\,\rho^{\,n-2k} .
\end{equation}
A pair $(n, m)$ represents a specific ``mode'' corresponding to $Z_n^m$.
These modes are defined only for values satisfying:
\begin{equation}
    n \ge 0, \quad \abs{m} \le n, \quad \text{$n - m$ is even} .
\end{equation}
The valid sequence of modes is thus \( (n, m) = (0, 0), (1, 1), (1, -1), (2, 0), \ldots \).

Zernike polynomials allow for the representation of arbitrary functions on a unit disk through a linear combination of basis polynomials:
\begin{equation}
f(\rho, \phi) = \sum_{(n, m)} a_{n,m} Z_n^m(\rho, \phi) ,
\end{equation}
where $a_{n,m}$ is the coefficient for the corresponding mode.
In our implementation, we fit the optical path differences (OPD) of the wavefront samples to the Zernike basis using the least-squares method.
To map the radial coordinate $\rho$ to the unit domain $[0, 1]$, we normalize the physical coordinates $(x, y)$ of the pupil samples by the exit pupil radius $\XPRad$.
We employ modes up to $n \le \OurZernN$, resulting in a total of 136 modes, which provides sufficient precision for our application.

\subsection{Handling Off-Axis Points}

\begin{figure*}[t]
  \centering
  \includegraphics[width=\linewidth]{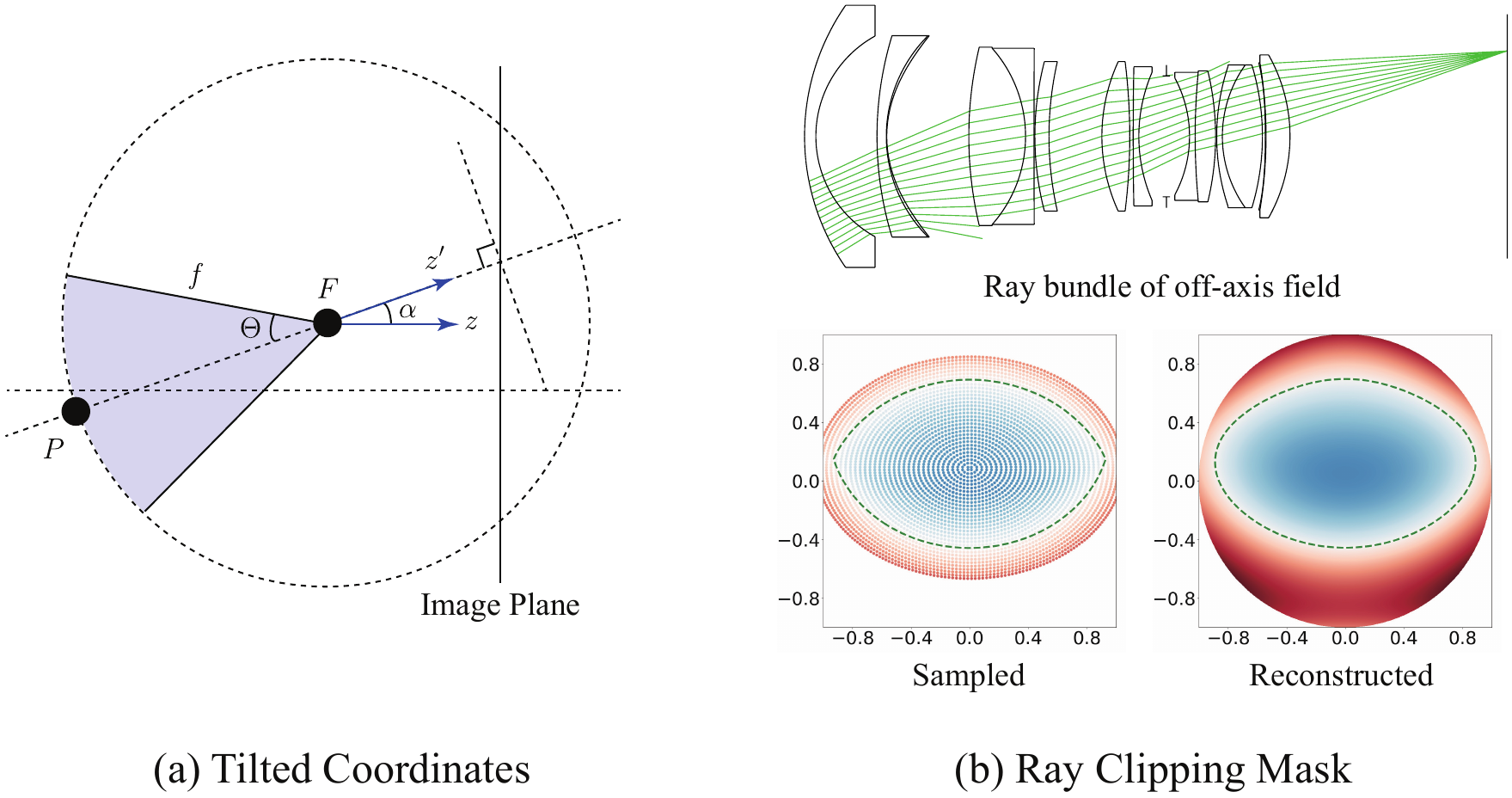}
  \caption{
  A figure illustrating our off-axis handling in the Debye formulation.
  (a) The coordinate system rotated by an angle $\alpha$ so that the focal point $\FocPt$ lies on the new optical axis.
  The image plane is now tilted with respect to the new coordinate system.
  (b) Sampling and reconstructing ray clipping values for complex pupil shapes.
  Blue denotes negative values (valid) and red denotes positive values (invalid); green curves denote the zero level set (mask boundary).
  }
  \label{fig:supp_debye_offaxis}
\end{figure*}

\providecommand{\TiltTheta}{\theta_0}
\providecommand{\TiltPhi}{\phi_0}
\providecommand{\kperp}{\vec{k}_{\perp}}
\providecommand{\RotMat}{R_T}
\providecommand{\NATheta}{\Theta}

The Debye formulation discussed thus far is strictly valid for on-axis focal points.
In practice, we also require PSFs for off-axis points, which result in off-axis focal points.
To address this, we define a local coordinate system that aligns the off-axis focal point $\FocPt$ with a new optical axis $z'$, as illustrated in \cref{fig:supp_debye_offaxis}-(a).
Within this rotated coordinate system, we apply the on-axis Debye formulation.

However, we cannot directly use the Debye CZT since the observation plane is no longer perpendicular to the new axis $z'$.\footnote{
	Note that this is not a limitation of the Debye formulation itself.
	Direct numerical integration can evaluate points on the untilted physical plane as well, although it is still computationally prohibitive.
}
To address this, we adopt the approach proposed by \cite{caiDirectCalculationTightly2019}.
It reformulates the Debye integral (\cref{eqn:supp_debye_integral}) with a tilted observation plane into a Fourier transform.
\cite{caiDirectCalculationTightly2019} first sets the rotated coordinates:
\begin{gather}
	\RotMat = {\setlength{\arraycolsep}{8pt}\begin{bmatrix} 
		\cos \TiltTheta \cos \TiltPhi & \cos \TiltTheta \sin \TiltPhi & -\sin \TiltTheta \\
		-\sin \TiltPhi & \cos \TiltPhi & 0 \\
		\sin \TiltTheta \cos \TiltPhi & \sin \TiltTheta \sin \TiltPhi & \cos \TiltTheta
	\end{bmatrix}} , \\
	\begin{bmatrix} u \\ v \\ w \end{bmatrix} = \RotMat \begin{bmatrix} x \\ y \\ z \end{bmatrix} , \quad
	\begin{bmatrix} k_u \\ k_v \\ k_w \end{bmatrix} = \RotMat \begin{bmatrix} k_x \\ k_y \\ k_z \end{bmatrix} ,
\end{gather}
where \( (\TiltTheta, \TiltPhi) \) are the polar and azimuthal angles of the new axis $z'$, $(u, v, w)$ are the new coordinates on the wavefront, and \( (k_u, k_v, k_w) \) is the corresponding new wave vector.\footnote{Note that the original matrix has $-\sin \TiltPhi$ instead of $-\sin \TiltTheta$ in the $(1,3)$ entry of Eq.~(7) of \cite{caiDirectCalculationTightly2019}. We believe this is a typo.}
Note that this easily reduces to the untilted Fourier form if we put \( (\TiltTheta, \TiltPhi) = (0, 0) \).
Next, \cite{caiDirectCalculationTightly2019} defines two integral domains\footnote{Note that Eq.~(9) in \cite{caiDirectCalculationTightly2019} contains $k$ instead of $k^2$ in the third row. As in the previous case, we believe this is a typo.}:
\begin{equation}
	\mathcal{D}_{\pm} (\TiltTheta, \TiltPhi) = \left\{ (k_u, k_v) \middle| 
	\begin{gathered}
		k_x(k_u, k_v, \pm \abs{k_w})^2 + k_y(k_u, k_v, \pm \abs{k_w})^2 < (k \sin \NATheta)^2 \\
		k \cos \NATheta < k_z(k_u, k_v, \pm \abs{k_w}) < k \\
		k_u^2 + k_v^2 < k^2
	\end{gathered}
	\right\} .
\end{equation}
Finally, \cite{caiDirectCalculationTightly2019} derives the Fourier form:
\begin{equation}
	\begin{bmatrix} E_x (\vec{x}) \\ E_y (\vec{x}) \\ E_z (\vec{x}) \end{bmatrix} = 
	\dfrac{C}{k} \mathcal{F}^{-1} \left[
		\left(
			\chi_+ (\kperp) e^{\Imag \abs{k_w} w} \begin{bmatrix} A_x \\ A_y \\ A_z \end{bmatrix}
			+ \chi_- (\kperp) e^{-\Imag \abs{k_w} w} \begin{bmatrix} A_x \\ A_y \\ A_z \end{bmatrix}
		\right) / \abs{k_w}
	\right] ,
\end{equation}
where $\chi_+$, $\chi_-$ denote membership functions of $\mathcal{D}_+$ and $\mathcal{D}_-$, respectively.
The Fourier transform is with respect to $(k_u, k_v)$, so we need to sample a grid on $(k_u, k_v)$ instead of $(k_x, k_y)$.
For detailed description, please refer to \cite{caiDirectCalculationTightly2019}.

Although the formulation is more complicated, it has the same computational structure as the untilted Debye CZT.
The only difference is the replacement of the inner term of the Fourier transform.
We did not additionally investigate the exact sampling criteria, but we found that the strengthened sampling criteria for the untilted case worked as well.

The remaining issue is the pupil shape.
Off-axis fields often cause complicated pupil shapes due to asymmetric light behavior and vignetting by clear apertures.
However, the Debye formulation originally assumes a perfect disk pupil.
To tackle this, we introduce a \emph{ray clipping mask} to simulate realistic pupil shapes.
Inspired by \cite{tehApertureAwareLensDesign2024}, we additionally evaluate the implicit function, dubbed a ``ray clipping function''.
It decides whether the ray is blocked by a clear aperture during ray tracing.
A ray is valid if and only if the corresponding clipping function value is negative, following a popular convention.
\cref{fig:supp_debye_offaxis}-(b) illustrates the overall process.
Like the phase values, we fit the sampled clipping function values to Zernike polynomials.
We found that high Zernike modes may introduce unstable extrapolation, resulting in incorrect masks.
Thus, we used modes up to only $n = 5$ with a total of $21$ modes.
This low-order representation has lower expressive power, which makes the reconstructed mask less accurate, as shown in \cref{fig:supp_debye_offaxis}-(b).
However, it is relatively robust against such erroneous extrapolation, which is often more critical.

The above techniques enable us to use the Debye CZT for off-axis fields as well.
Nevertheless, dealing with severe off-axis fields is still unstable due to the fundamental limitation of the Debye approximation and the risk of inaccurate ray clipping masks.
Therefore, we restrict the maximum field to $0.7$ for synthesis.

\subsection{Examples of Generated PSFs}

\begin{figure}[t]
	\centering
	\includegraphics[width=\linewidth]{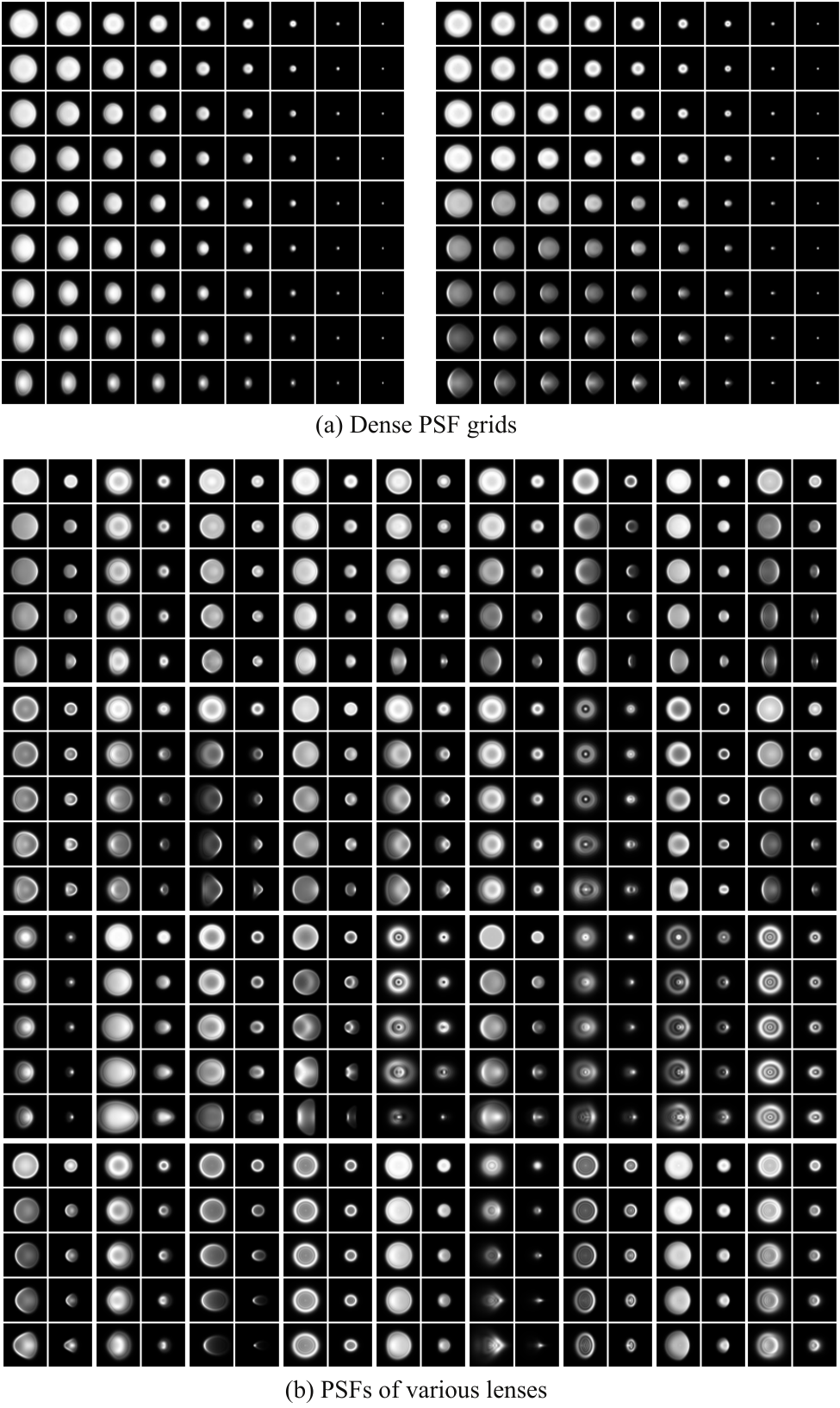}
	\caption{
    PSFs of various lenses, depths, and fields.
    Each grid corresponds to a distinct lens, where the row and column correspond to the field and depth respectively.
	}
	\label{fig:supp_psfs}
\end{figure}

\cref{fig:supp_psfs} illustrates various PSFs computed using the Debye CZT method for several compound lenses under different defocus conditions.
While modern photographic lenses are meticulously designed to minimize aberrations, the residual aberrations can still significantly influence the final PSF texture.
Additionally, for severe off-axis fields, vignetting often produces PSF shapes that resemble the cat's-eye bokeh effect.
These variations suggest that using datasets based on simplified PSF models or a limited selection of cameras may result in inherent domain gaps.

\subsection{Comparison with Zemax}

\begin{figure}[t]
	\centering
	\includegraphics[width=\linewidth]{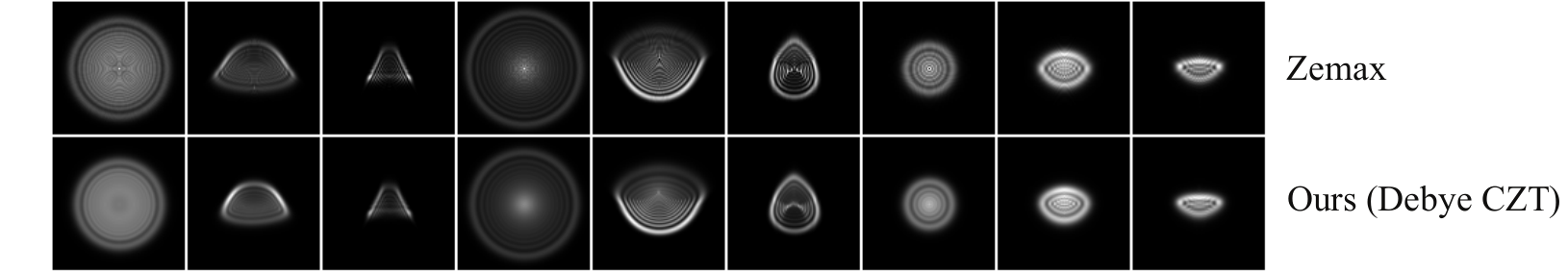}
	\caption{
		PSFs at three field positions for three lenses computed by Zemax and our method.
	}
	\label{fig:supp_zemax_val}
\end{figure}

To check the accuracy of the Debye CZT, we compared our results with PSFs computed by ``Huygens PSF'' of Zemax OpticStudio, which is often used as an accurate reference.
Unlike the Debye CZT, we manually choose the pupil sampling grid for Zemax.
\cref{fig:supp_zemax_val} shows PSFs of the same lenses and settings computed by two methods.
Our results are not fully accurate due to limitations such as the Debye approximation and imperfect ray clipping masks.
However, we believe that this accuracy is acceptable for our task, which aims to generate PSFs at scale rather than accurate manipulation of a specific system.

\subsection{Comparison with Other Methods}

We observe that the Debye formulation has received relatively little attention in computer vision and graphics, whereas several prior works have used the Huygens' principle~\cite{chenOpticalAberrationsCorrection2021, luoCorrectingOpticalAberration2024, hoDifferentiableWaveOptics2025, mullerExaminingImpactOptical2026} and Fraunhofer (FFT-based) PSF models~\cite{shahTiDyPSFsComputationalImaging2023, mullerClassificationRobustnessCommon2023, chenPhysicsInformedBlurLearning2025}.
However, the Debye CZT approach is superior for our purposes in several key aspects.
In this section, we compare our formulation with these two alternative methods.

\subsubsection{Huygens' Principle.}

The Huygens' principle is often described as the Rayleigh--Sommerfeld integral~\cite{marathayUsualApproximationUsed2004, goodman2017introduction}:
\begin{equation}\label{eqn:supp_rsi}
    U(\vec{x}) = \frac{1}{\Imag\lambda} \iint U_0(\vec{u}) \frac{ \exp(\Imag k \norm{ \vec{x} - \vec{u} } ) }{ \norm{\vec{x} - \vec{u}}^2 } \, [ \vec{n}(\vec{u}) \cdot (\vec{x} - \vec{u}) ] \, \dd S(\vec{u}) ,
\end{equation}
where $\vec{n}$ denotes the normal vector at point $\vec{u}$ on the surface.
Under the assumptions of scalar diffraction, this is considered one of the most accurate models for practical applications.
It can be computed directly by sampling the wave field $U_0(\vec{u})$ and numerically evaluating the integral for each observation point $\vec{x}$.

However, when evaluating an $M \times M$ viewport using $N \times N$ pupil samples, the time complexity is $\mathcal{O}(N^2 M^2)$.
This high cost stems from the nature of direct numerical integration, which requires computations for every pair of pupil and output grid points.
Furthermore, establishing robust sampling criteria for evaluating PSFs with large defocus across diverse compound lenses using the Huygens' principle is non-trivial.
This makes it difficult to determine a sampling density that reliably prevents aliasing without being computationally excessive.
Even if robust criteria were established, the resulting computation would be significantly slower than our approach, as analyzed in Sec.~5.3 of the main paper.
In conclusion, the Huygens' principle is less practical than the Debye CZT for large-scale generation of defocused PSFs.

\subsubsection{Fraunhofer PSF.}

The method commonly referred to as the ``FFT PSF'' has been widely adopted in prior works utilizing optical models~\cite{shahTiDyPSFsComputationalImaging2023, mullerClassificationRobustnessCommon2023, chenPhysicsInformedBlurLearning2025}. 
This approach models the PSF as a direct Fourier transform:
\begin{equation}\label{eqn:supp_fftpsf}
    E(x, y, z) \propto \mathcal{F}(P(x, y)) ,
\end{equation}
where $P$ is a pupil function, which represents wavefront aberrations.
Since the term ``FFT PSF'' technically describes a numerical implementation, we refer to it as the ``Fraunhofer PSF'' to emphasize its structural similarity to Fraunhofer diffraction.
Our analysis shows that this model can be viewed as a paraxial approximation of the Debye formulation.
While previous literature~\cite{leuteneggerFastFocusField2006, braat2008assessment, huEfficientFullpathOptical2020} has discussed this relationship, we provide a detailed derivation here for clarity.

Assume the numerical aperture (NA) is sufficiently small to satisfy the paraxial regime.
Under this condition, the spherical cap in the Debye formulation becomes nearly planar.
Since $k_x$ and $k_y$ are only defined within the disk domain \( k_x^2 + k_y^2 < (k \sin\Theta)^2 \), they are much smaller than $ k_z = \sqrt{k^2 - (k_x^2 + k_y^2)} $.
Applying a first-order Taylor expansion $ \sqrt{1 - t} \approx 1 - t/2 $, we can approximate:
\begin{equation}
    k_z \approx k - \frac{k_x^2 + k_y^2}{2 k} .
\end{equation}
Let $(\xi, \eta)$ denote the Cartesian coordinates on the exit pupil corresponding to the angular coordinates $(\theta, \phi)$ on the sphere, such that $\xi = - f \cos\phi \sin\theta$ and $\eta = - f \sin\phi \sin\theta$.
The wave vector components can then be rewritten as $k_x = k \xi / f$ and $k_y = k \eta / f$.
In the paraxial limit, $\theta$ is small, yielding $ \cos\theta \approx 1 $.
Substituting these into \cref{eqn:supp_debye_integral}, we obtain:
\begin{equation}\label{eqn:supp_debye_parax}
    \OutWave(x, y, z)
    \approx
    -\frac{\Imag}{\Wvl \FocLen} e^{\Imag k z}
    \iint \InWave(\xi, \eta) e^{ - \frac{\Imag k}{2 \FocLen^2} (\xi^2 + \eta^2) z } e^{-\frac{\Imag k}{\FocLen} (x \xi + y \eta)} \, \dd \xi \dd \eta .
\end{equation}
This expression is equivalent to \cref{eqn:supp_fftpsf}, with the addition of an inner quadratic phase factor with $z$ representing defocus.

Consequently, the Debye formulation can be understood as a more general and accurate version of the Fraunhofer PSF.
While it requires additional terms, such as the $1/\cos\theta$ apodization factor and exact $k_z$ mapping, these do not incur significant computational overhead.
We therefore regard the Debye CZT as a refined alternative to the Fraunhofer PSF, providing enhanced accuracy with affordable additional cost.

\section{Detailed Description of Lens Filtering}
\label{sec:supp_lens_designs}

In this section, we describe the detailed criteria used to filter lenses and the sampling process of optical parameters, which are omitted from the main paper.
As explained in the main paper, we first globally filter the lenses collected from the lens database to eliminate inappropriate types of lenses.
After that, we additionally filter lenses for each patch synthesis to restrict the number of samples and the size of defocus blur.

\subsection{Global Filtering}

\providecommand{\ImgRad}{I}
\providecommand{\TTL}{T}
\providecommand{\MaxPhaseDiff}{\Delta \Phi}

Because of the large scale of the lens database, it is difficult to manually choose appropriate lenses.
Thus, we establish a simple heuristic to classify the type of lens using only a few basic lens properties.
First, we check the F-number of a lens.
If it is larger than 8, we regard it as a non-photographic lens and eliminate it.
Next, we use two quantities: the image radius $\ImgRad$ and the total track length $\TTL$.
$\TTL$ is defined as the distance from the first surface to the focal plane (where the sensor is placed), which determines the physical length of the lens.
We classify lenses into two categories: interchangeable-lens cameras and smartphones.
If \( \ImgRad \ge \qty{10}{\milli\meter} \) and \( \TTL \ge \qty{25}{\milli\meter} \), we assume that the lens is for an interchangeable-lens camera.
If \( \ImgRad \in [2.5, 8]\si{\milli\meter} \) and \( \TTL \le \qty{15}{\milli\meter} \), we assume that the lens is for a smartphone camera.
We eliminate lenses that belong to neither category.

After filtering based on this classification, we additionally filter lenses for stable PSF generation.
We first eliminate lenses with $\NA \ge 0.6$.
This is because such lenses may exhibit strong vectorial effects and require high sampling numbers for the Debye CZT.
Moreover, such a large NA is unusual for photographic lenses\footnote{
This condition could be included in the previous classification together with the initial F-number restriction.
However, we intended to emphasize the issue of PSF computation rather than classification.
}.
Next, we measure the maximum phase difference $\MaxPhaseDiff$ between the wavefronts formed by an on-axis point and an off-axis point (at the edge of the field of view).
If $\MaxPhaseDiff$ is large, the lens would have severe aberrations.
This is unusual for well-designed modern photographic lenses, and it may also make PSF computation unstable.
Hence, we eliminate lenses with $\MaxPhaseDiff \ge 20\pi$ for either the on-axis or off-axis points.
Additionally, we found that there are occasionally incorrect design files that cannot form focus properly.
To address this, we eliminate lenses whose root-mean-square error (RMSE) of the spot points of the on-axis rays is larger than $0.01 \ImgRad$.

\subsection{Filtering and Sampling Per Patch Synthesis}

\providecommand{\DObj}{d_\mathrm{obj}}
\providecommand{\DImg}{d_\mathrm{img}}
\providecommand{\DSen}{d_\mathrm{sen}}
\providecommand{\DptMin}{d_\mathrm{min}}
\providecommand{\DptMax}{d_\mathrm{max}}
\providecommand{\CoCRad}{\mathrm{CoC}}
\providecommand{\CocRadMax}{\CoCRad_\mathrm{max}}
\providecommand{\CocRadLimit}{\CoCRad_\mathrm{limit}}
\providecommand{\PixSize}{\Delta x}
\providecommand{\SamNumMax}{N_\mathrm{max}}
\providecommand{\SamNumLimit}{N_\mathrm{limit}}

After the previous filtering, we can expect that the remaining lenses in the collection are appropriate.
However, combinations of a lens, a focusing distance, and a patch with multiple depths may still be inappropriate.
One potential issue is excessively large defocus, which is more likely to occur for lenses with larger apertures.
Another issue is that the Debye CZT may require an excessive sampling number according to the sampling condition, which also occurs more frequently for larger apertures.
To achieve a stable synthesis pipeline without manual tuning, we need to prevent such cases through per-patch filtering.

Our strategy is to assume the focusing distance that leads to the worst case for each lens and exclude all lenses that violate any constraint under that worst case.
After that, all focusing distances are safe to use, allowing us to sample a focusing distance without additional constraints.
Next, we explain the two constraints and filtering criteria in detail.

\subsubsection{Restriction on Defocus Size.}
\begin{figure}[t]
  \centering
  \includegraphics[width=0.8\linewidth]{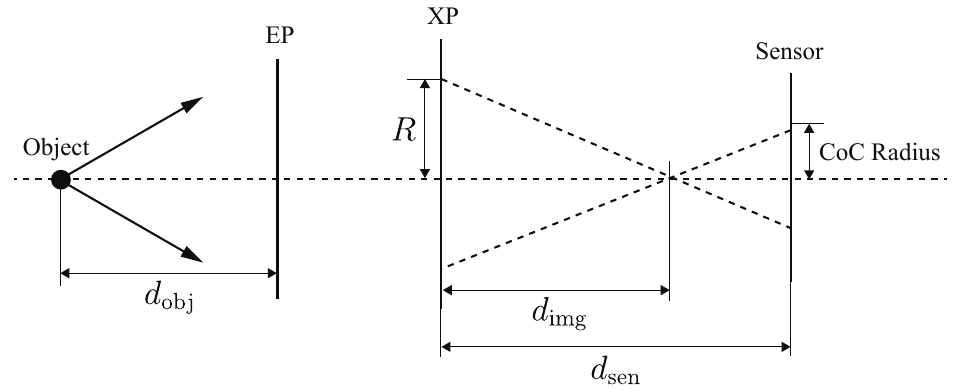}
  \caption{
  An illustration of the circle of confusion (CoC).
  The object distance $\DObj$ is measured from the entrance pupil (EP) plane.
  Rays from the object pass through the lens and converge to a point at $\DImg$, while the sensor is placed at $\DSen$.
  These distances are measured from the exit pupil (XP) plane.
  The rays intersecting the sensor form a CoC region with radius \( \XPRad \abs{(\DSen - \DImg) / \DImg} \).
  }
  \label{fig:supp_coc}
\end{figure}

We need to prevent excessively large PSFs that exceed a fixed kernel size.
Although it is difficult to determine the exact size of arbitrary PSFs, we can estimate it using the circle of confusion (CoC), defined under paraxial optics.
Consider a lens with exit pupil radius $\XPRad$.
If the sensor position is $\DSen$ and the object distance is $\DObj$, the CoC radius is:
\begin{equation}\label{eqn:supp_coc}
    \CoCRad (\DObj) = \XPRad \, \abs*{\frac{\DSen - \DImg(\DObj)}{\DImg(\DObj)}} ,
\end{equation}
where $\DImg(\DObj)$ denotes the conjugate image position of $\DObj$, as illustrated in \cref{fig:supp_coc}.
Both $\DSen$ and $\DImg(\DObj)$ can be calculated by the paraxial lens equation $1/f = 1/s + 1/s'$.
For a given sensor position and a multi-layered patch with depths $d_1, \ldots, d_n$, the CoC radius of each depth layer is $\CoCRad(d_1), \ldots, \CoCRad(d_n)$.
However, we note that the maximum value among them is always either $\CoCRad(\DptMin)$ or $\CoCRad(\DptMax)$, where $\DptMin$ and $\DptMax$ denote the minimum and maximum patch depths, respectively.
Hence, the maximum CoC radius is:
\begin{equation}
    \CocRadMax = \max \{ \CoCRad(\DptMin), \CoCRad(\DptMax) \} .
\end{equation}
Now, we restrict $\CocRadMax$ to be not greater than a threshold $\CocRadLimit$ by excluding all lenses that violate this condition.
We set:
\begin{equation}
    \CocRadLimit = \frac{1}{4} \KerSize \PixSize
\end{equation}
where $\KerSize$ is the kernel size (set to $128$ in our experiments) and $\PixSize$ denotes the pixel size.
This strictly restricts the CoC circle to lie within a box with $1/2$ size of the kernel viewport.
However, due to inaccuracies in paraxial geometric optics and complex aberrations that enlarge PSF shapes, we found this to be a reasonable constraint for keeping PSFs within the viewport while also allowing sizes that are not too small.

\subsubsection{Restriction on Sampling Number.}

In the main paper, the required sampling number for the Debye CZT was stated.
It depends on the defocus displacement $z$.
We also estimate it using paraxial optics.
For the PSF at depth $\DObj$, we estimate the sampling number as:
\begin{equation}
    N (\DObj) = 4 N_\mathrm{inf} (\DObj) = \frac{16 \NA^2}{\sqrt{n_t^2 - \NA^2}} \frac{\abs{\DSen - \DImg(\DObj)}}{\Wvl} .
\end{equation}
This also overestimates the sampling number by a factor of two to compensate for the gap between paraxial optics and real optics.
Analogous to the CoC restriction, $N (\DObj)$ attains its maximum either at $\DObj = \DptMin$ or at $\DObj = \DptMax$, yielding:
\begin{equation}
    \SamNumMax = \max \{ N (\DptMin), N (\DptMax) \} .
\end{equation}
We use the constraint \( \SamNumMax \le \SamNumLimit \) and exclude all lenses that violate it.
In our experiments, we set $\SamNumLimit = \num{1536}$.

\subsection{Diversity of Lens Collection}
\label{subsec:supp_lens_diversity}

\begin{figure*}[t]
	\centering
	\includegraphics{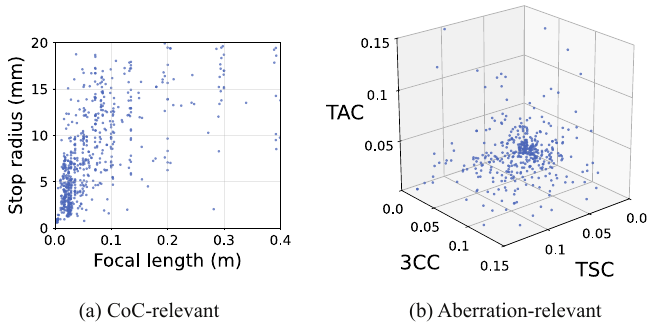}
	\caption{
		(a) Plot of distribution of focal length and stop radius.
		They are relevant to overall CoC size.
		(b) Plot of distribution of three values: TSC, 3CC, and TAC.
		They describe the strength of spherical, coma, and astigmatism aberrations.
	}
	\label{fig:supp_lens_diversity}
\end{figure*}

For our goal, the lens collection should provide diversity in defocus properties.
To validate that, we inspect several optical quantities.

First, we focus on two values: focal length and stop radius.
They decide the tendency of circle of confusion (CoC) size.
The CoC radius of a thin-lens model is formulated as~\cite{ruanAIFNetAllinFocusImage2021}:
\begin{equation}\label{eqn:supp_coc_thinlens}
	c = R \abs*{\frac{S_2 - S_1}{S_2} \cdot \frac{f}{S_1 - f}} ,
\end{equation}
where $f$ is the focal length, $R$ is the aperture radius, $S_1$ and $S_2$ are the focus and object distances, respectively.\footnote{
	This is a simplified version of \cref{eqn:supp_coc} for thin-lens models.
}
In practice, $S_1$ and $S_2$ are much larger than $f$, allowing us to replace $S_1 - f$ with $S_1$.
Then the CoC radius is approximately proportional to $f$ times $R$ for fixed $S_1$, $S_2$.
\cref{fig:supp_lens_diversity}-(a) shows the distribution of $f$ and $R$.
As the points are broadly scattered, $f R$ (area of rectangle formed by origin and point) varies across lenses, leading to various defocus sizes.

Next, we examine three values: TSC, 3CC, and TAC~\cite{smith2008modern}.
They are relevant to the strength of third-order spherical, coma, and astigmatism aberrations respectively.
\cref{fig:supp_lens_diversity}-(b) shows the distribution of three values.
As the points are scattered, we can expect diverse aberrations.

\section{Imaging Pipeline}
\label{sec:supp_synthesis}

\begin{figure}[t!]
  \centering
  \includegraphics[width=\linewidth]{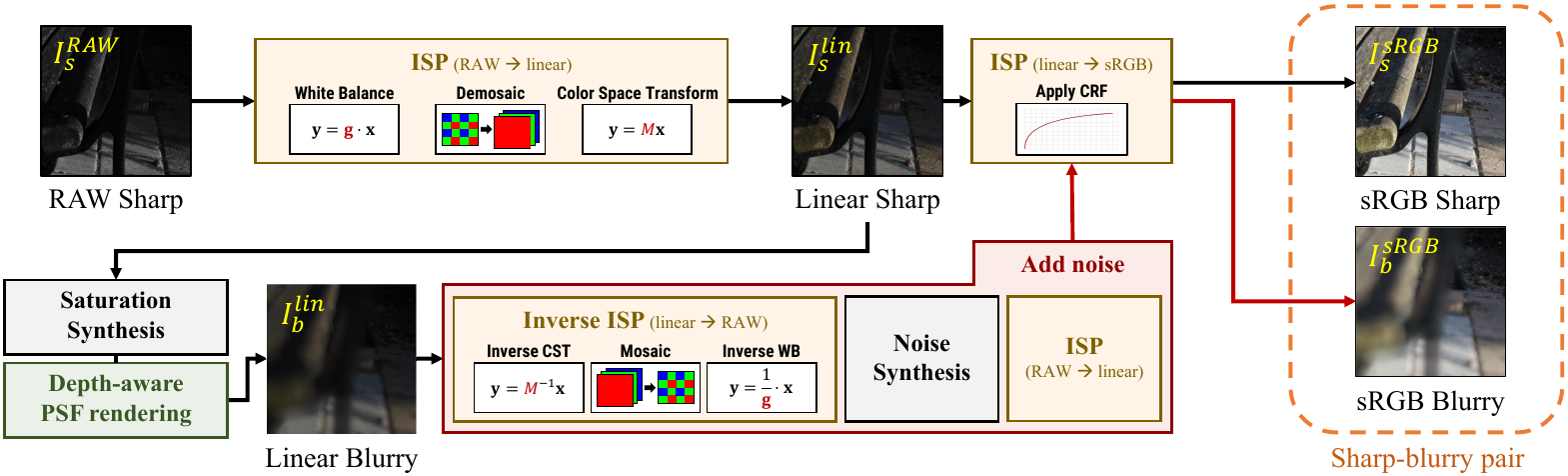}
  \caption{
  Overview of the imaging pipeline.
  $g$ denotes the multiplication factor used for white balancing, which is the reciprocal of the neutral shot value.
  $M$ denotes a linear transformation given by the product of the color correction matrix and the color space conversion matrix.
  }
  \label{fig:supp_isp}
\end{figure}

In this section, we describe the imaging pipeline used for our blur synthesis in detail.
We followed ISP models similar to \cite{abdelhamedHighQualityDenoisingDataset2018, rimRealisticBlurSynthesis2022}.
At a high level, the ISP consists of two stages: (1) RAW space to linear space, and (2) linear space to sRGB space.

\subsection{Overall Imaging Pipeline}

\cref{fig:supp_isp} illustrates the overall pipeline.
A RAW source image first passes through normalization with black and white levels and is converted into values within $[0, 1]$.
Next, it passes through the white balance stage, which divides each color channel by the neutral shot values provided in the metadata.
Then, we perform demosaicing to convert it into a continuous color map in camera RGB space.
Finally, we apply color correction and color space conversion by matrix multiplications, yielding the linear-space image.

In the linear space, we synthesize saturation (explained in \cref{subsec:supp_other_phenomena}) and defocus blur.
Then, we convert it back to RAW space by applying the inverse of the previous process.
After that, we synthesize noise (explained in \cref{subsec:supp_other_phenomena}) in RAW space.
As a result, we obtain the final blurred image in RAW space.
It then passes through the full forward ISP to be converted to sRGB space.
It first passes through the same process from RAW to linear space.
To convert it to sRGB space, we apply the camera response function (CRF) used in \cite{rimRealisticBlurSynthesis2022}.

\subsection{Synthesis of Other Phenomena}
\label{subsec:supp_other_phenomena}

\providecommand{\BetaRef}[1]{\beta_{#1,\textrm{ref}}}

While we focus on defocus blur, there are other physical phenomena that should also be considered.
We consider two additional phenomena, noise and saturation, which significantly affect the blur synthesis results.

\subsubsection{Noise Synthesis.}

Due to the nature of digital sensors, the sensor observation (RAW) contains noise.
A common approach to modeling the noise is to use a combination of Poisson noise and Gaussian noise~\cite{foiPracticalPoissonianGaussianNoise2008}.
The Poisson noise models signal-dependent photon shot noise, while the Gaussian noise models other signal-independent noise.
Similar to \cite{rimRealisticBlurSynthesis2022}, we model the total noise synthesis process as:
\begin{equation}
    I' = \beta_1 \mathcal{P}\left[ \frac{I}{\beta_1} \right] + \mathcal{N}(0, \beta_2) ,
\end{equation}
where $I$ is the input RAW, $\mathcal{P}$ and $\mathcal{N}$ denote the Poisson and Gaussian distributions, respectively, and $\beta_1, \beta_2$ are coefficients.
Following \cite{rimRealisticBlurSynthesis2022}, we sample $\beta_i$ from \( \mathcal{U} ( 0.5 \BetaRef{i}, 1.5 \BetaRef{i} ) \) for each $i = 1, 2$, rather than using fixed values, to provide diversity.
We set $\BetaRef{1} = \BetaRef{2} = \SI{1e-5}{}$.
This is much smaller than the noise strength used in prior work~\cite{rimRealisticBlurSynthesis2022}.
However, we found that stronger noise often decreases deblurring performance, which led us to keep these smaller values.

\subsubsection{Saturation Synthesis.}
\begin{figure}[t]
  \centering
  \includegraphics[width=0.8\linewidth]{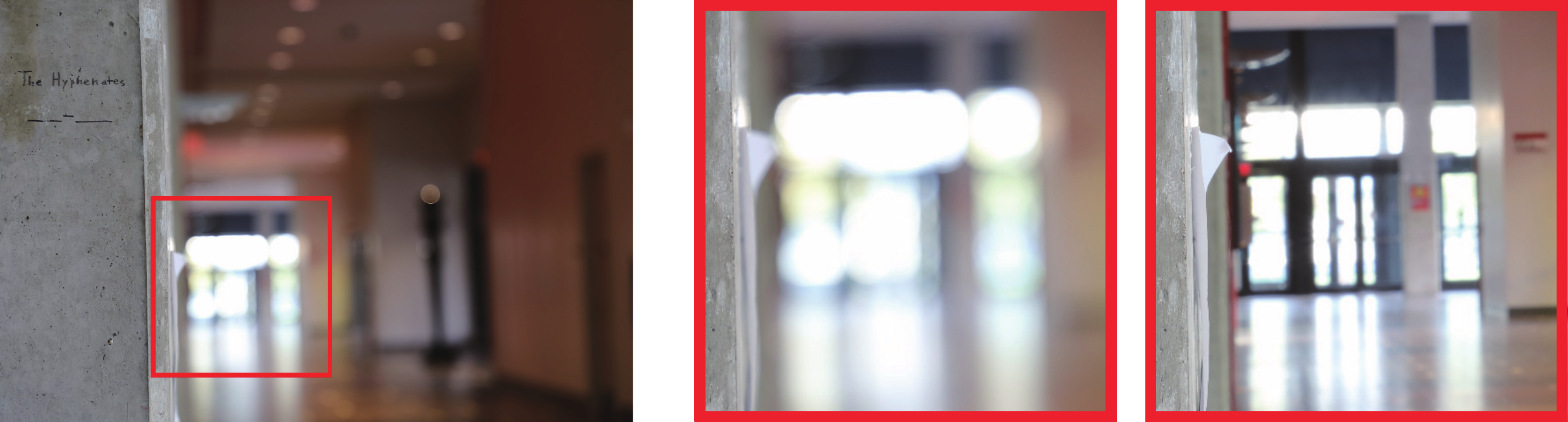}
  \caption{
  An example of defocus blur with saturation.
  Saturated pixels contain excess energy beyond the sensor limit, causing bright regions to expand after blurring.
  }
  \label{fig:supp_saturation}
\end{figure}

Because the dynamic range is limited, there may be saturated pixels whose actual energy is larger than the maximum pixel intensity.
If objects containing saturated pixels are blurred, they would spread more intensity depending on the amount of saturation.
In defocused images, this often appears as expanded bright regions, as shown in \cref{fig:supp_saturation}.
Hence, we synthesize saturation before applying blur in linear space.
To model this, we calculate the saturation mask by:
\begin{equation}
    M = \operatorname{clip} \left[ 20 \min_{c \in \{ R, G, B \}} (I_c - 0.95) \right] .
\end{equation}
Unlike a prior work~\cite{rimRealisticBlurSynthesis2022}, we use soft thresholding and a channel-independent mask, as this prevents unsatisfactory artifacts in our imaging pipeline.
After obtaining the mask, we apply the synthesis by:
\begin{equation}
    I' = \operatorname{clip} (I + \alpha M) ,
\end{equation}
where $\alpha$ is the saturation power sampled from $\mathcal{U}(0, 4)$.

\section{Extended Experimental Results}
\label{sec:supp_other_deblurring}

We provide extensive experimental results that are not included in the main paper.
First, we include three additional deblurring models: Restormer~\cite{zamirRestormerEfficientTransformer2022}, INIKNet~\cite{quanSingleImageDefocus2023}, and NAFNet~\cite{chenSimpleBaselinesImage2022}.
As there are no official weights of NAFNet trained on DPDD, we train it ourselves following the training protocol for motion deblurring.
Next, we also include more full-reference metrics (DISTS~\cite{dingImageQualityAssessment2022}) and no-reference metrics (MANIQA~\cite{yangMANIQAMultiDimensionAttention2022} and CLIP-IQA~\cite{wangExploringCLIPAssessing2023}).
\cref{tab:supp_all_metrics} shows quantitative results of all models and metrics.
They show a similar trend to the results in the main paper, supporting the consistent effect of our dataset across different deblurring models.

\begin{figure}[t]
  \centering
  \includegraphics[width=\linewidth]{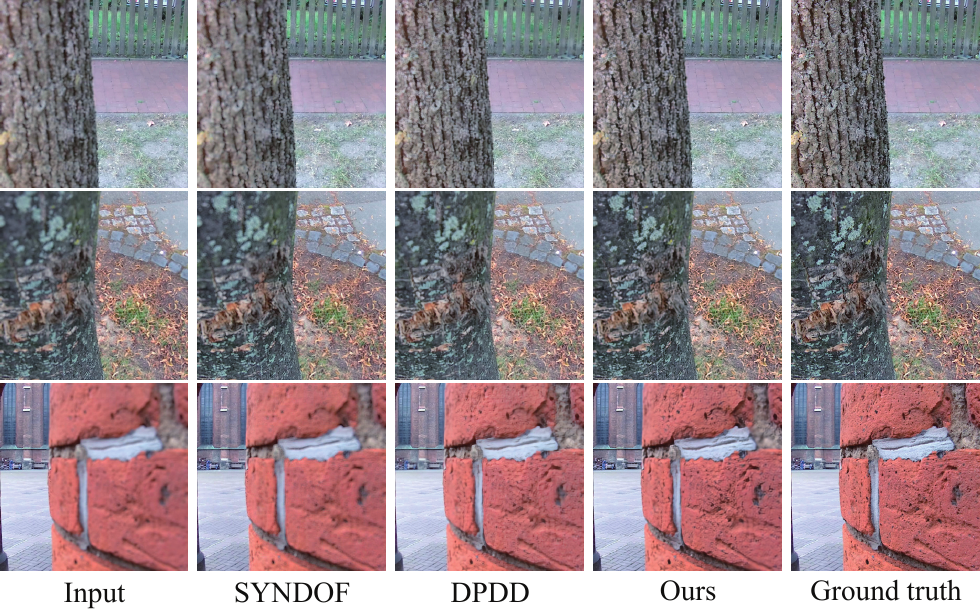}
  \caption{
  Qualitative comparisons on the RTF dataset.
  }
  \label{fig:supp_qualitative_extra_rtf}
\end{figure}

\begin{figure}[t]
  \centering
  \includegraphics[width=\linewidth]{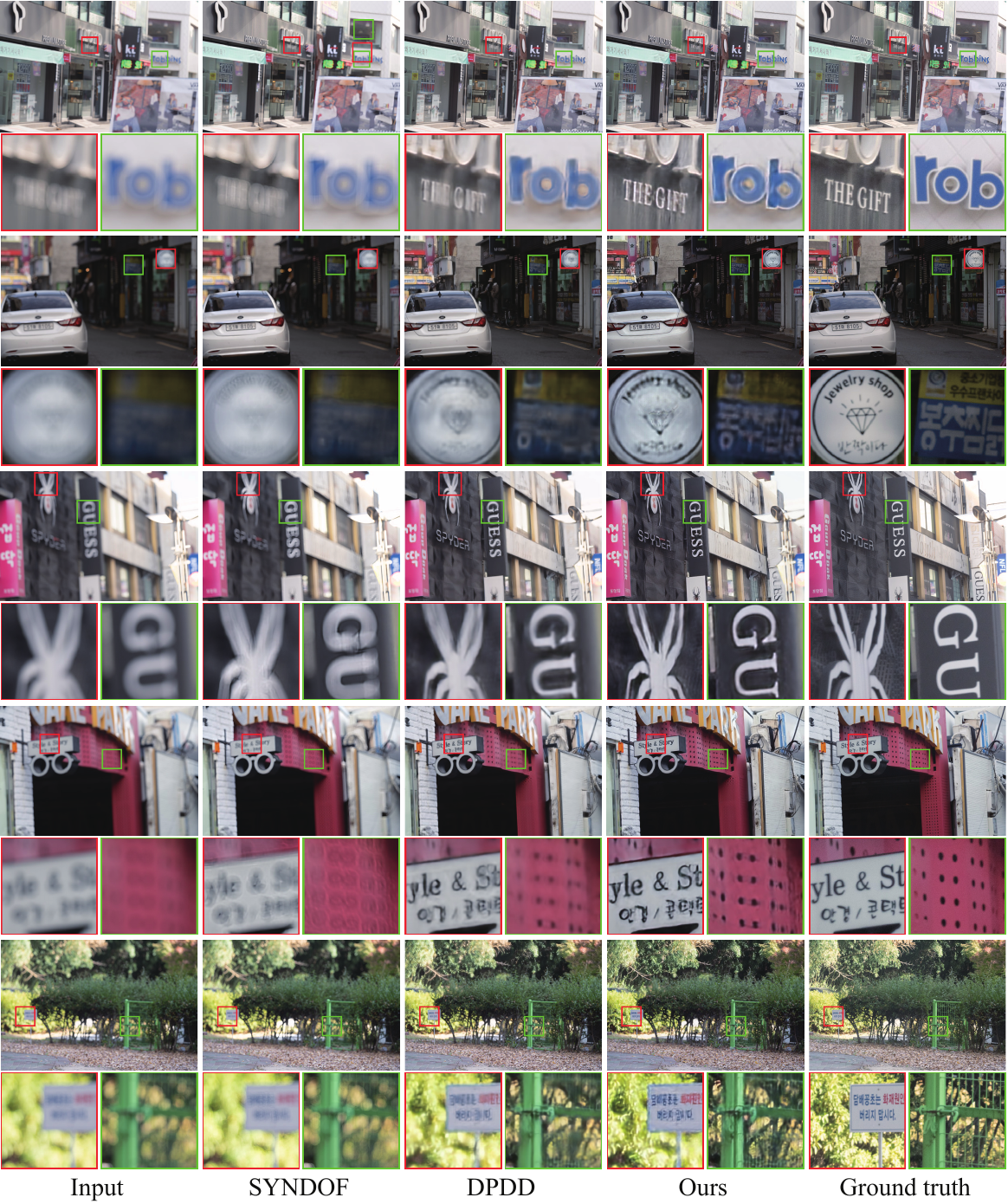}
  \caption{
  Additional qualitative comparisons on the RealDOF dataset.
  }
  \label{fig:supp_qualitative_extra}
\end{figure}

\begin{figure}[t]
  \centering
  \includegraphics[width=0.95\linewidth]{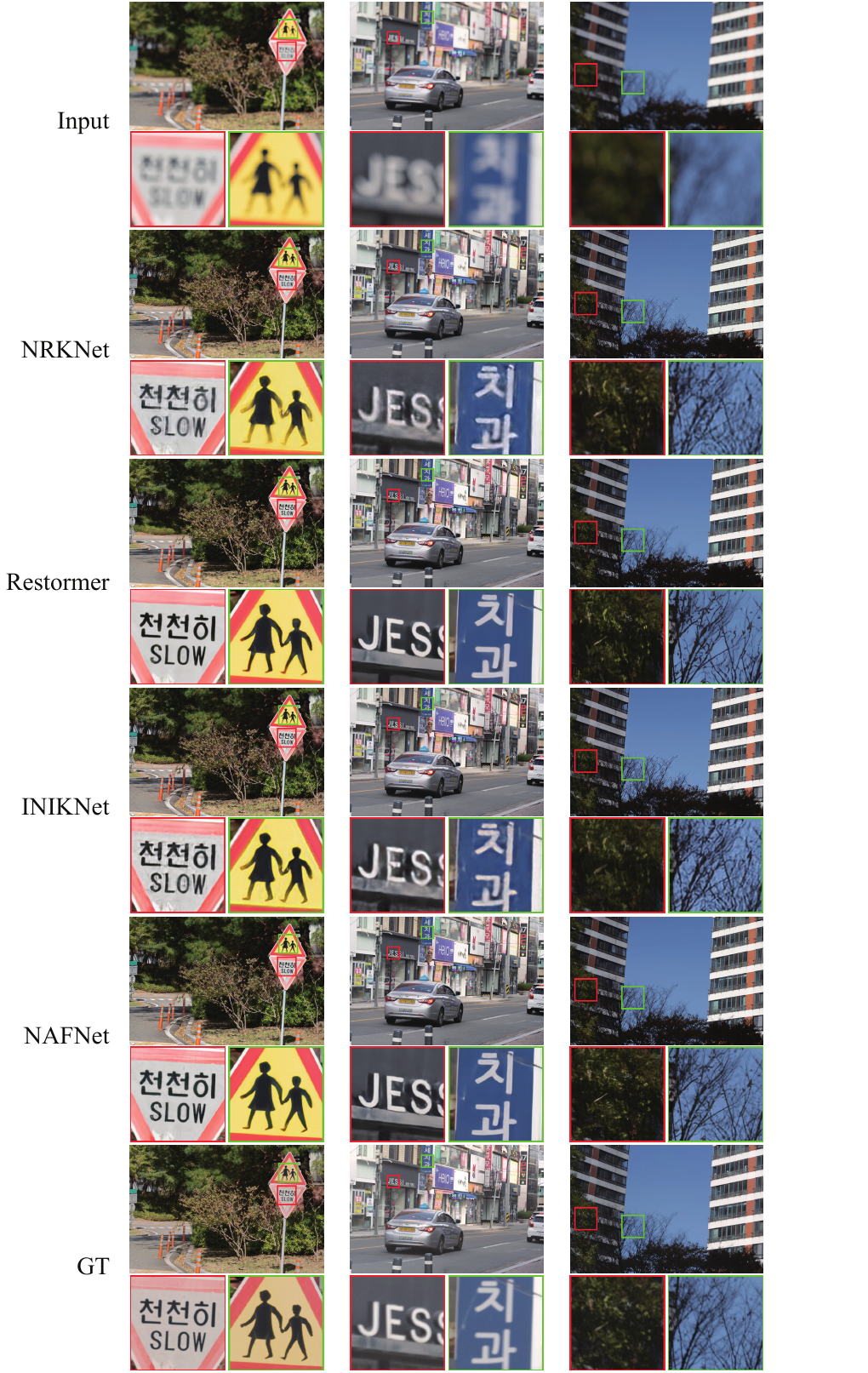}
  \caption{
  Qualitative comparisons on the same examples of the main paper's qualitative results (Fig.~3), across different deblurring models trained on CLDefocus.
  }
  \label{fig:supp_qualitative_across_models}
\end{figure}

Moreover, we provide additional qualitative results.
First, we provide additional results of NRKNet models in \cref{fig:supp_qualitative_extra} and \cref{fig:supp_qualitative_extra_rtf} on the RealDOF and RTF, respectively.
Next, we provide results across deblurring models in \cref{fig:supp_qualitative_across_models} with the same examples of the main paper.
They show that not only NRKNet but also other models trained on CLDefocus work as well, proving the consistent effectiveness of CLDefocus.
Interestingly, the results of Restormer and NAFNet achieve better perceptual quality than that of NRKNet, and they even seem sharper than the ground-truth images with slight residual blur.

\section{Evaluation on Images Captured by a Smartphone}
\label{sec:supp_smartphone}

The optical design of smartphone cameras is constrained by physical requirements such as total thickness~\cite{blahnikSmartphoneImagingTechnology2021}.
Therefore, smartphone lenses typically consist of compact high-order aspheric surfaces, unlike the lenses of common interchangeable-lens cameras.
This makes the lens blur of smartphone cameras significantly different from that of other cameras, introducing a domain gap from real-captured datasets such as DPDD~\cite{abuolaimDefocusDeblurringUsing2020}.

In this section, we evaluate the generalization ability of our synthetic dataset and two other datasets: SYNDOF~\cite{leeDeepDefocusMap2019} and DPDD~\cite{abuolaimDefocusDeblurringUsing2020} on smartphone cameras.
For this, we captured 100 photos using the wide lens (default zoom) of a Samsung Galaxy S24.
Due to the difficulty of capturing sharp pairs, we only collected defocused images.
Since smartphone cameras usually have broad depth-of-field regions, significant defocus appears when capturing close objects (typically closer than $\qty{0.3}{\meter}$).
Thus, we considered common real-life scenarios involving close objects.
The photos were captured at a resolution of $4080 \times 3060$ and then downscaled by a factor of 2, yielding $2040 \times 1530$.

\begin{table}[t]
	\centering
	\setlength{\tabcolsep}{2pt}
	\caption{Quantitative comparisons on smartphone images. The \colorbox{red!30}{best}, \colorbox{orange!25}{second best}, and \colorbox{yellow!25}{third best} results are highlighted.}
	\label{tab:supp_smartphone}
	\begin{tabular}{lccccc}
		\toprule
		Train set & NIQE$\downarrow$ & MUSIQ$\uparrow$ & MANIQA$\uparrow$ & TOPIQ$\uparrow$ & CLIP-IQA$\uparrow$ \\
		\midrule
		SYNDOF & \cellcolor{yellow!25}4.267 & \cellcolor{yellow!25}45.680 & \cellcolor{red!30}0.280 & \cellcolor{yellow!25}0.397 & \cellcolor{yellow!25}0.383 \\
		DPDD & \cellcolor{orange!25}3.973 & \cellcolor{orange!25}46.016 & \cellcolor{yellow!25}0.274 & \cellcolor{orange!25}0.400 & \cellcolor{orange!25}0.396 \\
		Ours & \cellcolor{red!30}3.609 & \cellcolor{red!30}47.888 & \cellcolor{orange!25}0.278 & \cellcolor{red!30}0.406 & \cellcolor{red!30}0.423 \\
		\bottomrule
	\end{tabular}
\end{table}

\begin{figure}[t]
  \centering
  \includegraphics[width=\linewidth]{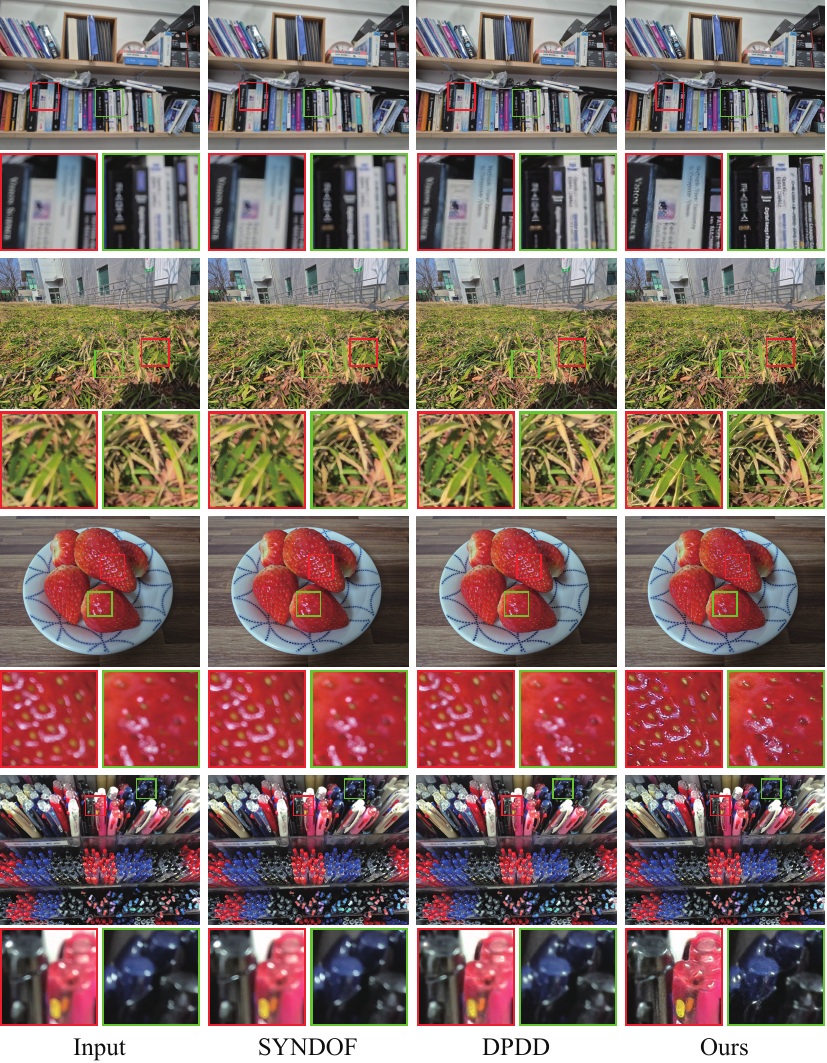}
  \caption{Qualitative comparisons on defocused images captured by a Samsung Galaxy S24. In this example, the SYNDOF-trained model almost failed to detect blur. The DPDD-trained model weakly reduced blur. The CLDefocus-trained achieved the strongest deblurred results.}
  \label{fig:supp_smartphone_qualitative}
\end{figure}

We evaluated NRKNet~\cite{quanNeumannNetworkRecursive2023} models trained respectively on SYNDOF, DPDD, and our CLDefocus dataset, identical to those used in the main paper.
The quantitative evaluation using no-reference metrics is reported in \cref{tab:supp_smartphone}.
Moreover, the qualitative results for a few images are shown in \cref{fig:supp_smartphone_qualitative}.
Both results show that the model trained on our dataset outperforms the ones trained on others, indicating stronger generalization ability.
In particular, the model trained on other datasets often fails to properly handle blur with complicated PSF shapes, as in the strawberry example in \cref{fig:supp_smartphone_qualitative}.
We presume that this is because the PSFs of smartphone cameras are significantly different from the lens domain of Gaussian blur and DPDD, making the model less aware of such types of blur.
We believe this further supports the effectiveness of our approach for realistic synthetic datasets.

\section{Downstream Vision Tasks}
\label{sec:supp_downstream}

We aim to check the effect of defocus deblurring on downstream vision tasks.
However, public datasets for vision tasks with real defocus blur are rarely found.
There exist datasets with various degradations including defocus blur~\cite{hendrycks2019robustness}, but they mostly rely on simplified synthesis.
There are a few real-captured datasets such as \cite{ljosaAnnotatedHighthroughputMicroscopy2012} for microscopes.
However, they do not fit our domain as we focus on ordinary photographic lenses.
Thus, instead of using existing datasets, we use the RealDOF dataset with pseudo ground-truth (GT) labels estimated from sharp GT images.
Using the same estimation models, we obtain the labels from blurred inputs and images restored by NRKNet trained on SYNDOF, DPDD, and our CLDefocus dataset.

\begin{figure}[t]
	\centering
	\includegraphics[width=\linewidth]{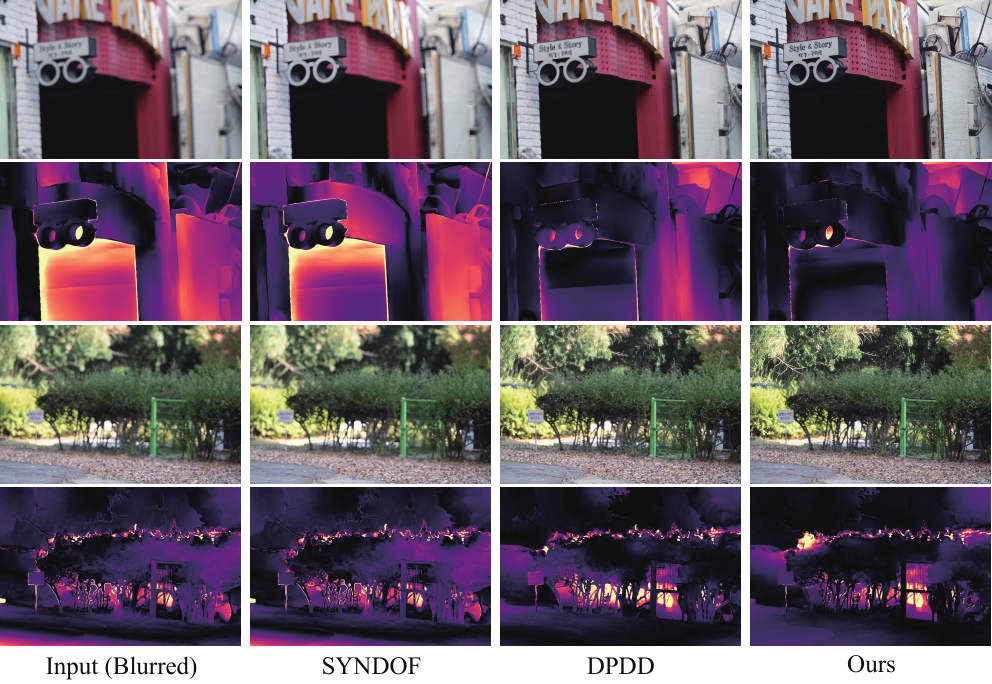}
	\caption{
		Qualitative results of depth estimation.
		The first and third rows are images, and the second and fourth rows are the error maps relative to the result on the sharp GT image (pseudo-GT).
	}
	\label{fig:supp_downstream_depth}
\end{figure}

We test two common vision tasks: monocular depth estimation and semantic segmentation.
For depth estimation, we use Depth Anything V2~\cite{yangDepthAnythingV22024}.
Since we cannot trust the absolute depth values of pseudo-GTs, we use a relative depth model and focus on how the structure is preserved rather than absolute metrics.
Following \cite{ranftlRobustMonocularDepth2022}, we evaluate the root-mean-squared-error (RMSE) in disparity space.
We use the RMSE both with alignment (to match scale and shift) and without alignment.
The former one is a natural choice for relative depth estimation.
The latter one is to check how the direct estimator output is affected by the defocus blur and deblurring, as the pseudo-GT is also estimated by the same model.
\cref{tab:supp_downstream_depth} and \cref{fig:supp_downstream_depth} show the quantitative and qualitative results.
The SYNDOF-trained model does not improve the performance from blurred inputs.
The DPDD-trained and CLDefocus-trained models significantly improve quality.
Moreover, the CLDefocus-trained model achieves slightly better metrics.

\begin{table}[t]
	\centering
	\setlength{\tabcolsep}{4pt}
	\caption{Quantitative result of depth estimation on the RealDOF dataset.}
	\label{tab:supp_downstream_depth}
	\begin{tabular}{lcccc}
		\toprule
		Train set & Input & SYNDOF & DPDD & Ours \\
		\midrule
		RMSE (aligned) & 0.330 & 0.331 & 0.248 & \textbf{0.246} \\
		RMSE (not aligned) & 0.630 & 0.587 & 0.360 & \textbf{0.344} \\
		\bottomrule
	\end{tabular}
\end{table}

\begin{figure}[t]
	\centering
	\includegraphics{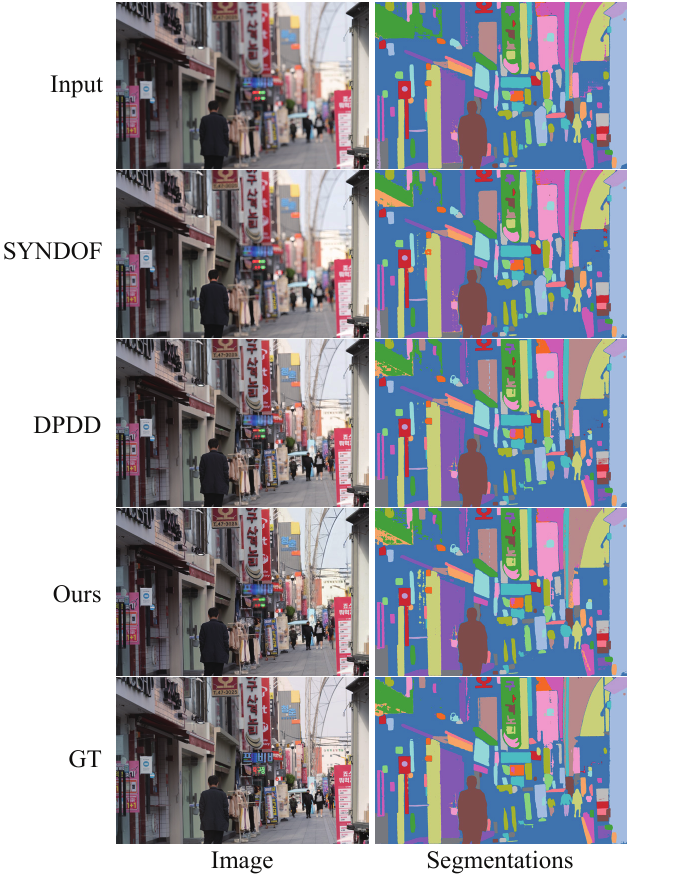}
	\caption{
		Qualitative results of semantic segmentation using SAM 2.
	}
	\label{fig:supp_downstream_segment}
\end{figure}

For semantic segmentation, we use Segment Anything Model 2 (SAM 2)~\cite{raviSAM2Segment2025}.
It takes prompts like point markers and masks and generates segmentation labels for each prompt without explicit classes.
We use the following protocol.
First, we generate mask prompts for a sharp GT image using the automatic mask generator provided by SAM 2.
Then we estimate segmentation labels of the GT image for each mask prompt using SAM 2.
Next, we estimate the input and restored images with the same prompts of the GT image to keep output consistency.
Finally, we evaluate the mean of intersection over union (IoU) per prompt for each image.
\cref{tab:supp_downstream_segment} and \cref{fig:supp_downstream_segment} show the quantitative and qualitative results.
They show a similar trend to depth estimation.
The SYNDOF-trained model shows poor performance, and the DPDD-trained and CLDefocus-trained models achieve improvement while the latter one records slightly better scores.

\begin{table}[t]
	\centering
	\setlength{\tabcolsep}{4pt}
	\caption{Quantitative result of semantic segmentation on the RealDOF dataset.}
	\label{tab:supp_downstream_segment}
	\begin{tabular}{lcccc}
		\toprule
		Train set & Input & SYNDOF & DPDD & Ours \\
		\midrule
		Mean IoU & 0.769 & 0.761 & 0.867 & \textbf{0.871} \\
		\bottomrule
	\end{tabular}
\end{table}

\newpage
\begin{table}[t]
\centering
\setlength{\tabcolsep}{2pt}
\renewcommand{\arraystretch}{1.2}
\caption{Quantitative comparisons across test sets, deblurring models, and training sets. The \colorbox{red!30}{best}, \colorbox{orange!25}{second best}, and \colorbox{yellow!25}{third best} results are highlighted.}
\label{tab:supp_all_metrics}
\resizebox{1\linewidth}{!}{
\begin{tabular}{lllccccccccc}
\toprule
Test set & Model & Train & PSNR$\uparrow$ & SSIM$\uparrow$ & LPIPS$\downarrow$ & DISTS$\downarrow$ & NIQE$\downarrow$ & MUSIQ$\uparrow$ & MANIQA$\uparrow$ & TOPIQ$\uparrow$ & CLIP-IQA$\uparrow$ \\
\midrule
\multirow{12}{*}{Ours} & \multirow{3}{*}{NRKNet} & SYNDOF & \cellcolor{yellow!25}27.58 & \cellcolor{yellow!25}0.790 & \cellcolor{yellow!25}0.296 & \cellcolor{yellow!25}0.209 & \cellcolor{yellow!25}12.048 & \cellcolor{yellow!25}36.904 & \cellcolor{yellow!25}0.252 & \cellcolor{yellow!25}0.316 & \cellcolor{yellow!25}0.276 \\
 &  & DPDD & \cellcolor{orange!25}28.27 & \cellcolor{orange!25}0.825 & \cellcolor{orange!25}0.263 & \cellcolor{orange!25}0.195 & \cellcolor{red!30}7.508 & \cellcolor{orange!25}38.961 & \cellcolor{orange!25}0.277 & \cellcolor{orange!25}0.345 & \cellcolor{orange!25}0.300 \\
 &  & Ours & \cellcolor{red!30}32.16 & \cellcolor{red!30}0.865 & \cellcolor{red!30}0.201 & \cellcolor{red!30}0.164 & \cellcolor{orange!25}7.724 & \cellcolor{red!30}39.850 & \cellcolor{red!30}0.287 & \cellcolor{red!30}0.348 & \cellcolor{red!30}0.331 \\
\cline{2-12}
 & \multirow{3}{*}{Restormer} & SYNDOF & \cellcolor{orange!25}29.31 & \cellcolor{yellow!25}0.799 & \cellcolor{yellow!25}0.331 & \cellcolor{yellow!25}0.226 & \cellcolor{yellow!25}8.782 & \cellcolor{yellow!25}29.693 & \cellcolor{yellow!25}0.251 & \cellcolor{yellow!25}0.269 & \cellcolor{yellow!25}0.271 \\
 &  & DPDD & \cellcolor{yellow!25}27.57 & \cellcolor{orange!25}0.829 & \cellcolor{orange!25}0.233 & \cellcolor{orange!25}0.184 & \cellcolor{red!30}7.629 & \cellcolor{orange!25}42.052 & \cellcolor{orange!25}0.293 & \cellcolor{orange!25}0.376 & \cellcolor{orange!25}0.318 \\
 &  & Ours & \cellcolor{red!30}35.26 & \cellcolor{red!30}0.910 & \cellcolor{red!30}0.125 & \cellcolor{red!30}0.121 & \cellcolor{orange!25}7.705 & \cellcolor{red!30}44.614 & \cellcolor{red!30}0.316 & \cellcolor{red!30}0.387 & \cellcolor{red!30}0.358 \\
\cline{2-12}
 & \multirow{3}{*}{INIKNet} & SYNDOF & \cellcolor{yellow!25}27.84 & \cellcolor{yellow!25}0.787 & \cellcolor{yellow!25}0.299 & \cellcolor{yellow!25}0.207 & \cellcolor{orange!25}7.203 & \cellcolor{red!30}40.799 & \cellcolor{orange!25}0.274 & \cellcolor{red!30}0.363 & \cellcolor{yellow!25}0.295 \\
 &  & DPDD & \cellcolor{orange!25}28.49 & \cellcolor{orange!25}0.829 & \cellcolor{orange!25}0.233 & \cellcolor{orange!25}0.182 & \cellcolor{red!30}6.953 & \cellcolor{yellow!25}39.743 & \cellcolor{yellow!25}0.272 & \cellcolor{orange!25}0.360 & \cellcolor{orange!25}0.308 \\
 &  & Ours & \cellcolor{red!30}32.55 & \cellcolor{red!30}0.869 & \cellcolor{red!30}0.170 & \cellcolor{red!30}0.152 & \cellcolor{yellow!25}7.619 & \cellcolor{orange!25}40.490 & \cellcolor{red!30}0.281 & \cellcolor{yellow!25}0.357 & \cellcolor{red!30}0.339 \\
\cline{2-12}
 & \multirow{3}{*}{NAFNet} & SYNDOF & \cellcolor{orange!25}29.05 & \cellcolor{yellow!25}0.799 & \cellcolor{yellow!25}0.330 & \cellcolor{yellow!25}0.226 & \cellcolor{yellow!25}8.637 & \cellcolor{yellow!25}31.596 & \cellcolor{yellow!25}0.261 & \cellcolor{yellow!25}0.289 & \cellcolor{yellow!25}0.286 \\
 &  & DPDD & \cellcolor{yellow!25}26.56 & \cellcolor{orange!25}0.808 & \cellcolor{orange!25}0.252 & \cellcolor{orange!25}0.195 & \cellcolor{red!30}7.113 & \cellcolor{orange!25}43.163 & \cellcolor{orange!25}0.299 & \cellcolor{red!30}0.390 & \cellcolor{orange!25}0.340 \\
 &  & Ours & \cellcolor{red!30}34.91 & \cellcolor{red!30}0.904 & \cellcolor{red!30}0.127 & \cellcolor{red!30}0.119 & \cellcolor{orange!25}7.633 & \cellcolor{red!30}44.528 & \cellcolor{red!30}0.318 & \cellcolor{orange!25}0.388 & \cellcolor{red!30}0.362 \\
\midrule
Test set & Model & Train & PSNR$\uparrow$ & SSIM$\uparrow$ & LPIPS$\downarrow$ & DISTS$\downarrow$ & NIQE$\downarrow$ & MUSIQ$\uparrow$ & MANIQA$\uparrow$ & TOPIQ$\uparrow$ & CLIP-IQA$\uparrow$ \\
\midrule
\multirow{12}{*}{DPDD} & \multirow{3}{*}{NRKNet} & SYNDOF & \cellcolor{yellow!25}23.77 & \cellcolor{yellow!25}0.743 & \cellcolor{yellow!25}0.315 & \cellcolor{yellow!25}0.157 & \cellcolor{yellow!25}5.067 & \cellcolor{yellow!25}55.819 & \cellcolor{orange!25}0.345 & \cellcolor{orange!25}0.493 & \cellcolor{red!30}0.446 \\
 &  & DPDD & \cellcolor{red!30}26.11 & \cellcolor{red!30}0.817 & \cellcolor{red!30}0.223 & \cellcolor{red!30}0.140 & \cellcolor{orange!25}4.746 & \cellcolor{red!30}59.473 & \cellcolor{red!30}0.358 & \cellcolor{red!30}0.507 & \cellcolor{yellow!25}0.406 \\
 &  & Ours & \cellcolor{orange!25}24.73 & \cellcolor{orange!25}0.776 & \cellcolor{orange!25}0.265 & \cellcolor{orange!25}0.146 & \cellcolor{red!30}4.657 & \cellcolor{orange!25}56.696 & \cellcolor{yellow!25}0.338 & \cellcolor{yellow!25}0.487 & \cellcolor{orange!25}0.408 \\
\cline{2-12}
 & \multirow{3}{*}{Restormer} & SYNDOF & \cellcolor{yellow!25}23.89 & \cellcolor{yellow!25}0.745 & \cellcolor{yellow!25}0.348 & \cellcolor{yellow!25}0.182 & \cellcolor{yellow!25}5.500 & \cellcolor{yellow!25}55.413 & \cellcolor{orange!25}0.354 & \cellcolor{orange!25}0.495 & \cellcolor{red!30}0.474 \\
 &  & DPDD & \cellcolor{red!30}25.98 & \cellcolor{red!30}0.822 & \cellcolor{red!30}0.176 & \cellcolor{red!30}0.120 & \cellcolor{red!30}4.985 & \cellcolor{red!30}61.779 & \cellcolor{red!30}0.389 & \cellcolor{red!30}0.539 & \cellcolor{orange!25}0.441 \\
 &  & Ours & \cellcolor{orange!25}24.91 & \cellcolor{orange!25}0.779 & \cellcolor{orange!25}0.277 & \cellcolor{orange!25}0.164 & \cellcolor{orange!25}5.326 & \cellcolor{orange!25}56.752 & \cellcolor{yellow!25}0.334 & \cellcolor{yellow!25}0.458 & \cellcolor{yellow!25}0.399 \\
\cline{2-12}
 & \multirow{3}{*}{INIKNet} & SYNDOF & \cellcolor{yellow!25}23.81 & \cellcolor{yellow!25}0.743 & \cellcolor{yellow!25}0.316 & \cellcolor{orange!25}0.158 & \cellcolor{orange!25}5.094 & \cellcolor{orange!25}56.147 & \cellcolor{red!30}0.351 & \cellcolor{orange!25}0.500 & \cellcolor{red!30}0.441 \\
 &  & DPDD & \cellcolor{red!30}26.10 & \cellcolor{red!30}0.818 & \cellcolor{red!30}0.184 & \cellcolor{red!30}0.123 & \cellcolor{red!30}4.584 & \cellcolor{red!30}59.305 & \cellcolor{orange!25}0.350 & \cellcolor{red!30}0.502 & \cellcolor{yellow!25}0.409 \\
 &  & Ours & \cellcolor{orange!25}24.57 & \cellcolor{orange!25}0.772 & \cellcolor{orange!25}0.273 & \cellcolor{yellow!25}0.161 & \cellcolor{yellow!25}5.161 & \cellcolor{yellow!25}55.854 & \cellcolor{yellow!25}0.317 & \cellcolor{yellow!25}0.428 & \cellcolor{orange!25}0.415 \\
\cline{2-12}
 & \multirow{3}{*}{NAFNet} & SYNDOF & \cellcolor{yellow!25}23.88 & \cellcolor{yellow!25}0.746 & \cellcolor{yellow!25}0.344 & \cellcolor{yellow!25}0.179 & \cellcolor{yellow!25}5.510 & \cellcolor{yellow!25}55.830 & \cellcolor{orange!25}0.359 & \cellcolor{orange!25}0.502 & \cellcolor{red!30}0.479 \\
 &  & DPDD & \cellcolor{red!30}25.60 & \cellcolor{red!30}0.814 & \cellcolor{red!30}0.184 & \cellcolor{red!30}0.119 & \cellcolor{red!30}4.736 & \cellcolor{red!30}61.869 & \cellcolor{red!30}0.390 & \cellcolor{red!30}0.544 & \cellcolor{orange!25}0.460 \\
 &  & Ours & \cellcolor{orange!25}25.16 & \cellcolor{orange!25}0.794 & \cellcolor{orange!25}0.228 & \cellcolor{orange!25}0.135 & \cellcolor{orange!25}4.758 & \cellcolor{orange!25}59.625 & \cellcolor{yellow!25}0.350 & \cellcolor{yellow!25}0.481 & \cellcolor{yellow!25}0.436 \\
\midrule
Test set & Model & Train & PSNR$\uparrow$ & SSIM$\uparrow$ & LPIPS$\downarrow$ & DISTS$\downarrow$ & NIQE$\downarrow$ & MUSIQ$\uparrow$ & MANIQA$\uparrow$ & TOPIQ$\uparrow$ & CLIP-IQA$\uparrow$ \\
\midrule
\multirow{12}{*}{RealDOF} & \multirow{3}{*}{NRKNet} & SYNDOF & \cellcolor{yellow!25}21.64 & \cellcolor{yellow!25}0.654 & \cellcolor{yellow!25}0.474 & \cellcolor{yellow!25}0.232 & \cellcolor{yellow!25}6.285 & \cellcolor{yellow!25}23.905 & \cellcolor{red!30}0.242 & \cellcolor{yellow!25}0.241 & \cellcolor{red!30}0.276 \\
 &  & DPDD & \cellcolor{red!30}25.03 & \cellcolor{red!30}0.771 & \cellcolor{orange!25}0.335 & \cellcolor{orange!25}0.182 & \cellcolor{orange!25}6.030 & \cellcolor{orange!25}31.350 & \cellcolor{yellow!25}0.219 & \cellcolor{orange!25}0.270 & \cellcolor{yellow!25}0.272 \\
 &  & Ours & \cellcolor{orange!25}24.74 & \cellcolor{orange!25}0.764 & \cellcolor{red!30}0.303 & \cellcolor{red!30}0.162 & \cellcolor{red!30}5.029 & \cellcolor{red!30}35.787 & \cellcolor{orange!25}0.226 & \cellcolor{red!30}0.303 & \cellcolor{orange!25}0.272 \\
\cline{2-12}
 & \multirow{3}{*}{Restormer} & SYNDOF & \cellcolor{yellow!25}22.31 & \cellcolor{yellow!25}0.666 & \cellcolor{yellow!25}0.521 & \cellcolor{yellow!25}0.284 & \cellcolor{yellow!25}7.743 & \cellcolor{yellow!25}24.454 & \cellcolor{orange!25}0.268 & \cellcolor{yellow!25}0.237 & \cellcolor{yellow!25}0.285 \\
 &  & DPDD & \cellcolor{red!30}25.08 & \cellcolor{red!30}0.784 & \cellcolor{orange!25}0.286 & \cellcolor{orange!25}0.157 & \cellcolor{orange!25}6.017 & \cellcolor{orange!25}37.492 & \cellcolor{yellow!25}0.242 & \cellcolor{orange!25}0.320 & \cellcolor{orange!25}0.290 \\
 &  & Ours & \cellcolor{orange!25}24.48 & \cellcolor{orange!25}0.776 & \cellcolor{red!30}0.277 & \cellcolor{red!30}0.150 & \cellcolor{red!30}5.274 & \cellcolor{red!30}48.173 & \cellcolor{red!30}0.288 & \cellcolor{red!30}0.391 & \cellcolor{red!30}0.342 \\
\cline{2-12}
 & \multirow{3}{*}{INIKNet} & SYNDOF & \cellcolor{yellow!25}22.06 & \cellcolor{yellow!25}0.659 & \cellcolor{yellow!25}0.486 & \cellcolor{yellow!25}0.243 & \cellcolor{yellow!25}6.804 & \cellcolor{yellow!25}24.607 & \cellcolor{red!30}0.254 & \cellcolor{yellow!25}0.242 & \cellcolor{yellow!25}0.271 \\
 &  & DPDD & \cellcolor{red!30}25.31 & \cellcolor{red!30}0.782 & \cellcolor{red!30}0.287 & \cellcolor{red!30}0.162 & \cellcolor{orange!25}5.791 & \cellcolor{orange!25}33.390 & \cellcolor{yellow!25}0.215 & \cellcolor{orange!25}0.281 & \cellcolor{orange!25}0.290 \\
 &  & Ours & \cellcolor{orange!25}24.24 & \cellcolor{orange!25}0.761 & \cellcolor{orange!25}0.293 & \cellcolor{orange!25}0.165 & \cellcolor{red!30}5.659 & \cellcolor{red!30}37.582 & \cellcolor{orange!25}0.237 & \cellcolor{red!30}0.314 & \cellcolor{red!30}0.348 \\
\cline{2-12}
 & \multirow{3}{*}{NAFNet} & SYNDOF & \cellcolor{yellow!25}22.24 & \cellcolor{yellow!25}0.665 & \cellcolor{yellow!25}0.515 & \cellcolor{yellow!25}0.276 & \cellcolor{yellow!25}7.441 & \cellcolor{yellow!25}25.939 & \cellcolor{orange!25}0.271 & \cellcolor{yellow!25}0.253 & \cellcolor{yellow!25}0.287 \\
 &  & DPDD & \cellcolor{orange!25}24.45 & \cellcolor{orange!25}0.765 & \cellcolor{orange!25}0.299 & \cellcolor{orange!25}0.161 & \cellcolor{orange!25}5.677 & \cellcolor{orange!25}36.032 & \cellcolor{yellow!25}0.229 & \cellcolor{orange!25}0.313 & \cellcolor{orange!25}0.295 \\
 &  & Ours & \cellcolor{red!30}24.46 & \cellcolor{red!30}0.767 & \cellcolor{red!30}0.273 & \cellcolor{red!30}0.138 & \cellcolor{red!30}4.780 & \cellcolor{red!30}49.537 & \cellcolor{red!30}0.290 & \cellcolor{red!30}0.395 & \cellcolor{red!30}0.357 \\
\midrule
Test set & Model & Train & PSNR$\uparrow$ & SSIM$\uparrow$ & LPIPS$\downarrow$ & DISTS$\downarrow$ & NIQE$\downarrow$ & MUSIQ$\uparrow$ & MANIQA$\uparrow$ & TOPIQ$\uparrow$ & CLIP-IQA$\uparrow$ \\
\midrule
\multirow{12}{*}{RTF} & \multirow{3}{*}{NRKNet} & SYNDOF & \cellcolor{yellow!25}24.55 & \cellcolor{yellow!25}0.743 & \cellcolor{yellow!25}0.261 & \cellcolor{yellow!25}0.120 & \cellcolor{red!30}3.855 & \cellcolor{yellow!25}57.279 & \cellcolor{yellow!25}0.349 & \cellcolor{red!30}0.539 & \cellcolor{orange!25}0.480 \\
 &  & DPDD & \cellcolor{orange!25}25.95 & \cellcolor{orange!25}0.833 & \cellcolor{red!30}0.207 & \cellcolor{red!30}0.112 & \cellcolor{orange!25}4.196 & \cellcolor{orange!25}58.822 & \cellcolor{orange!25}0.358 & \cellcolor{yellow!25}0.531 & \cellcolor{yellow!25}0.419 \\
 &  & Ours & \cellcolor{red!30}26.62 & \cellcolor{red!30}0.845 & \cellcolor{orange!25}0.242 & \cellcolor{orange!25}0.112 & \cellcolor{yellow!25}4.263 & \cellcolor{red!30}59.144 & \cellcolor{red!30}0.373 & \cellcolor{orange!25}0.533 & \cellcolor{red!30}0.519 \\
\cline{2-12}
 & \multirow{3}{*}{Restormer} & SYNDOF & \cellcolor{orange!25}24.45 & \cellcolor{yellow!25}0.734 & \cellcolor{yellow!25}0.274 & \cellcolor{yellow!25}0.122 & \cellcolor{red!30}3.824 & \cellcolor{yellow!25}56.532 & \cellcolor{yellow!25}0.355 & \cellcolor{yellow!25}0.545 & \cellcolor{orange!25}0.506 \\
 &  & DPDD & \cellcolor{yellow!25}24.24 & \cellcolor{orange!25}0.807 & \cellcolor{orange!25}0.205 & \cellcolor{red!30}0.110 & \cellcolor{yellow!25}3.968 & \cellcolor{orange!25}59.934 & \cellcolor{orange!25}0.375 & \cellcolor{orange!25}0.553 & \cellcolor{yellow!25}0.426 \\
 &  & Ours & \cellcolor{red!30}26.90 & \cellcolor{red!30}0.866 & \cellcolor{red!30}0.169 & \cellcolor{orange!25}0.112 & \cellcolor{orange!25}3.957 & \cellcolor{red!30}63.486 & \cellcolor{red!30}0.415 & \cellcolor{red!30}0.587 & \cellcolor{red!30}0.529 \\
\cline{2-12}
 & \multirow{3}{*}{INIKNet} & SYNDOF & \cellcolor{yellow!25}24.70 & \cellcolor{yellow!25}0.748 & \cellcolor{yellow!25}0.267 & \cellcolor{yellow!25}0.116 & \cellcolor{red!30}3.709 & \cellcolor{yellow!25}57.783 & \cellcolor{orange!25}0.368 & \cellcolor{red!30}0.547 & \cellcolor{orange!25}0.491 \\
 &  & DPDD & \cellcolor{orange!25}25.48 & \cellcolor{orange!25}0.817 & \cellcolor{red!30}0.201 & \cellcolor{red!30}0.115 & \cellcolor{orange!25}4.039 & \cellcolor{orange!25}57.963 & \cellcolor{yellow!25}0.344 & \cellcolor{orange!25}0.522 & \cellcolor{yellow!25}0.422 \\
 &  & Ours & \cellcolor{red!30}26.72 & \cellcolor{red!30}0.856 & \cellcolor{orange!25}0.218 & \cellcolor{orange!25}0.116 & \cellcolor{yellow!25}4.523 & \cellcolor{red!30}60.784 & \cellcolor{red!30}0.372 & \cellcolor{yellow!25}0.518 & \cellcolor{red!30}0.501 \\
\cline{2-12}
 & \multirow{3}{*}{NAFNet} & SYNDOF & \cellcolor{orange!25}24.45 & \cellcolor{yellow!25}0.740 & \cellcolor{orange!25}0.220 & \cellcolor{red!30}0.109 & \cellcolor{red!30}3.542 & \cellcolor{orange!25}58.928 & \cellcolor{orange!25}0.380 & \cellcolor{orange!25}0.567 & \cellcolor{orange!25}0.494 \\
 &  & DPDD & \cellcolor{yellow!25}24.26 & \cellcolor{orange!25}0.798 & \cellcolor{yellow!25}0.237 & \cellcolor{yellow!25}0.120 & \cellcolor{yellow!25}4.139 & \cellcolor{yellow!25}58.678 & \cellcolor{yellow!25}0.373 & \cellcolor{yellow!25}0.539 & \cellcolor{yellow!25}0.428 \\
 &  & Ours & \cellcolor{red!30}25.85 & \cellcolor{red!30}0.856 & \cellcolor{red!30}0.180 & \cellcolor{orange!25}0.113 & \cellcolor{orange!25}3.870 & \cellcolor{red!30}61.535 & \cellcolor{red!30}0.395 & \cellcolor{red!30}0.570 & \cellcolor{red!30}0.507 \\
\bottomrule
\end{tabular}}
\end{table}

\end{document}